%% file: arxiv.tex
\documentclass[sigconf]{acmart}
\usepackage{algorithm}
\usepackage{algorithmic}
\definecolor{darkpurple}{RGB}{102, 51, 153}
\usepackage{enumitem}
\usepackage{threeparttable}
\usepackage{color}
\usepackage{multirow}
\usepackage{array}
\usepackage{amsmath}
\usepackage{pifont}
\newcommand\blfootnote[1]{%
  \begingroup
  \renewcommand\thefootnote{}\footnote{#1}%
  \addtocounter{footnote}{-1}%
  \endgroup
}
\usepackage{balance}

\usepackage{tikz}

\usepackage[edges]{forest}
\definecolor{hidden-draw}{RGB}{0,0,0}
\definecolor{hidden-pink}{rgb}{0.98, 0.94, 0.75}
\definecolor{level0}{rgb}{0.67, 0.88, 0.69}
\definecolor{level1}{rgb}{0.98, 0.92, 0.84}
\definecolor{level2}{rgb}{0.8, 0.8, 1.0}
\definecolor{level3}{rgb}{1.0, 0.71, 0.76}
\definecolor{level4}{rgb}{0.49, 0.99, 0.0}

\newcommand{\definitionautorefname}{Definition}

\newcommand*{\shortautoref}[1]{%
  \begingroup
    \def\sectionautorefname{Sec.}%
    \def\subsectionautorefname{Sec.}%
    \def\figureautorefname{Fig.}%
    \def\tableautorefname{Tab.}%
    \def\equationautorefname{Eq.}%
    \def\subfigureautorefname{Fig.}%
    \def\definitionautorefname{Def.}%
    \autoref{#1}%
  \endgroup
}

\definecolor{lawngreen}{rgb}{0.49, 0.99, 0.0}
\definecolor{pink}{rgb}{1, 0, 0.5}
\definecolor{airforce}{rgb}{0.36, 0.54, 0.66}

\hypersetup{
    linkbordercolor=airforce,
    urlbordercolor=pink,
    citebordercolor=green,
}

\copyrightyear{2024}
\acmYear{2024}
\setcopyright{acmlicensed}\acmConference[KDD '24]{Proceedings of the 30th ACM SIGKDD Conference on Knowledge Discovery and Data Mining}{August 25--29, 2024}{Barcelona, Spain}
\acmBooktitle{Proceedings of the 30th ACM SIGKDD Conference on Knowledge Discovery and Data Mining (KDD '24), August 25--29, 2024, Barcelona, Spain}
\acmDOI{10.1145/3637528.3671451}
\acmISBN{979-8-4007-0490-1/24/08}
\AtBeginDocument{%
  \providecommand\BibTeX{{%
    \normalfont B\kern-0.5em{\scshape i\kern-0.25em b}\kern-0.8em\TeX}}}

\begin{document}


\title[Foundation Models for Time Series Analysis: A Tutorial and Survey]{Foundation Models for Time Series Analysis: \\ A Tutorial and Survey}




\author{Yuxuan Liang}
\affiliation{%
  \institution{The Hong Kong University of Science and Technology (Guangzhou)
  }
  \city{} 
  \state{}
  \country{}
}
\email{yuxliang@outlook.com}

\author{Haomin Wen}
\affiliation{%
  \institution{Beijing Jiao Tong University \& HKUST(GZ)
  }
  \city{} 
  \state{}
  \country{}
}
\email{wenhaomin@bjtu.edu.cn}

\author{Yuqi Nie}
\affiliation{%
  \institution{Princeton University
  }
  \city{} 
  \state{}
  \country{}
}
\email{ynie@princeton.edu}

\author{Yushan Jiang}
\affiliation{%
  \institution{University of Connecticut
  }
  \city{} 
  \state{}
  \country{}
}
\email{yushan.jiang@uconn.edu}

\author{Ming Jin}
\affiliation{%
  \institution{Monash University
  }
  \city{} 
  \state{}
  \country{}
}
\email{ming.jin@monash.edu}

\author{Dongjin Song}
\affiliation{%
  \institution{University of Connecticut
  }
  \city{} 
  \state{}
  \country{}
}
\email{dongjin.song@uconn.edu}

\author{Shirui Pan}
\affiliation{%
  \institution{Griffith University
  }
  \city{} 
  \state{}
  \country{}
}
\email{s.pan@griffith.edu.au}

\author{Qingsong Wen}
\affiliation{%
  \institution{Squirrel AI, USA
  }
  \city{} 
  \state{}
  \country{}
}
\email{qingsongedu@gmail.com}

\renewcommand{\shortauthors}{Yuxuan Liang et al.}

\begin{abstract}
Time series analysis stands as a focal point within the data mining community, serving as a cornerstone for extracting valuable insights crucial to a myriad of real-world applications. Recent advances in Foundation Models (FMs) have fundamentally reshaped the paradigm of model design for time series analysis, boosting various downstream tasks in practice. These innovative approaches often leverage pre-trained or fine-tuned FMs to harness generalized knowledge tailored for time series analysis. This survey aims to furnish a comprehensive and up-to-date overview of FMs for time series analysis. While prior surveys have predominantly focused on either application or pipeline aspects of FMs in time series analysis, they have often lacked an in-depth understanding of the underlying mechanisms that elucidate why and how FMs benefit time series analysis. To address this gap, our survey adopts a methodology-centric classification, delineating various pivotal elements of time-series FMs, including model architectures, pre-training techniques, adaptation methods, and data modalities. Overall, this survey serves to consolidate the latest advancements in FMs pertinent to time series analysis, accentuating their theoretical underpinnings, recent strides in development, and avenues for future exploration.

\blfootnote{Q. Wen is the corresponding author.}
\end{abstract}


\maketitle

%
\section{Introduction} 

\begin{figure}[!t]
    \centering
    \includegraphics[width=0.95 \linewidth]{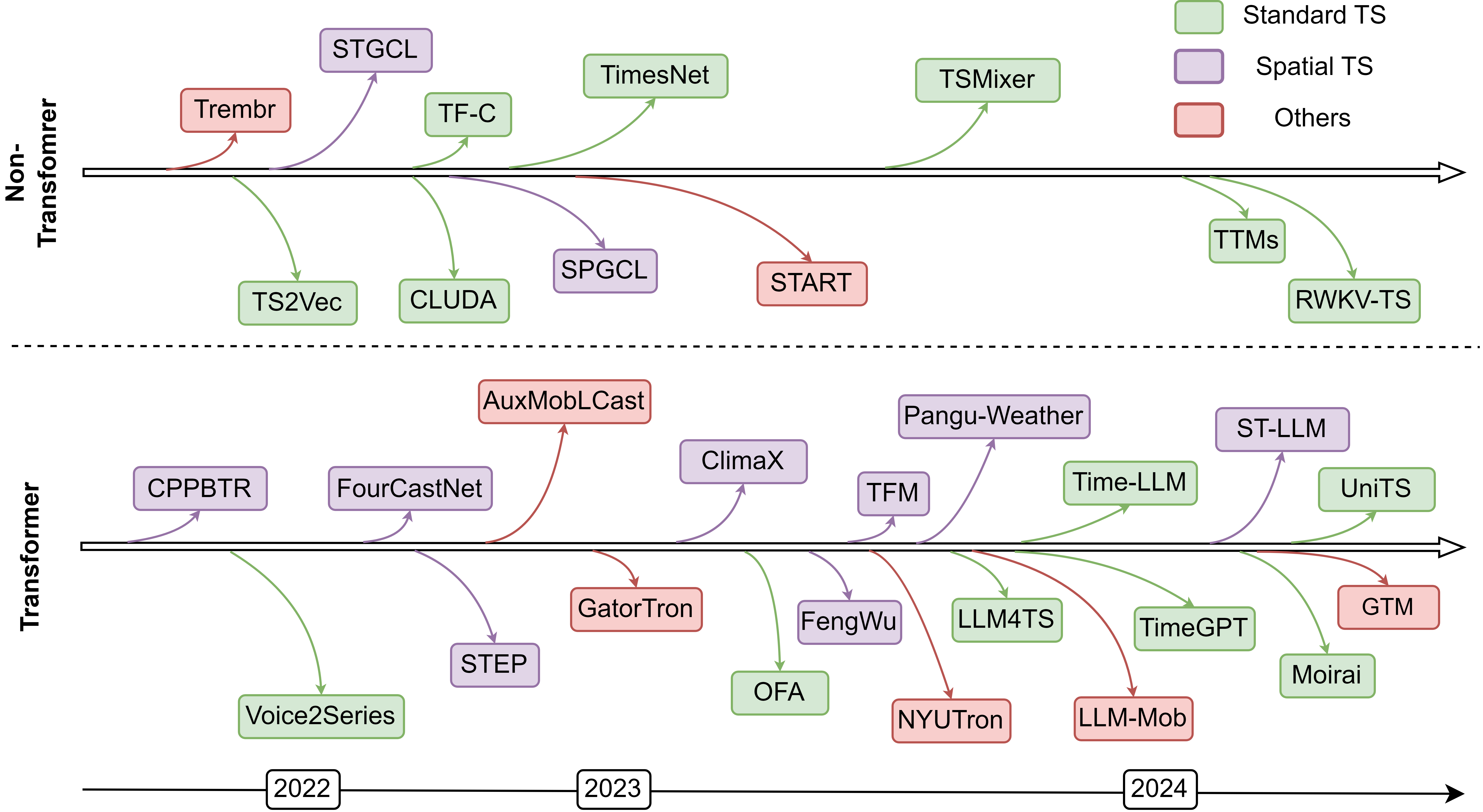}
    \caption{Roadmaps of representative TSFMs.}
    \label{fig:roadmap}
\end{figure}
Time series data are characterized by their sequential order and temporal dependencies, encapsulating valuable information about the dynamics of diverse systems and processes \cite{zhang2023self, jiang2024empowering, wang2024deep}. Various time series data (e.g., stock price, traffic flow, electricity) present unique challenges and opportunities for computational analysis, each requiring tailored approaches to effectively capture their inherent properties. The analysis and understanding of time series data is an important piece of data mining, facilitating crucial insights and decisions in many domains~\cite{wen2022robust,jin2023survey}, including finance \cite{yu2023temporal, chen2023chatgpt}, healthcare \cite{liu2023large, li2023frozen}, cloud computing~\cite{zhang2021cloudrca}, environments \cite{pathak2022fourcastnet, chen2023prompt}, energy~\cite{zhu2023energy}, and urban computing \cite{wang2023building, shao2022pre}.

In recent years, the advancements of deep learning, especially the transformer-based models~\cite{vaswani2017transformer}, have revolutionized the field of time series analysis~\cite{wen2022transformers}. Following the pioneering work~\cite{li2019enhancing}, there has been a surge in research interest exploring the application of transformers for time series analysis~\cite{zhou2021informer,zhou2022fedformer,xu2021anomaly}.
The motivation behind deep learning and transformers lies in their ability to automatically learn comprehensive representations from raw data, thus capturing complex nonlinear relationships and temporal dependencies without the need for manual feature engineering. Such capability leads to significant performance improvements compared with traditional statistical methods across numerous time series applications. Foundation models (FMs), such as large language models (LLMs) in natural language processing (NLP) \cite{zhao2023survey} and advanced models in computer vision (CV) \cite{awais2023foundational}, have emerged as powerful paradigms capable of achieving state-of-the-art performances in their respective fields. The success of these FMs can be attributed to their ability to \emph{leverage vast amounts of data to cultivate general-purpose representations}, subsequently fine-tuning them, or even deploying them directly in a zero-shot manner to excel across a diverse spectrum of downstream tasks. This approach not only economizes on the need for task-specific model development but also encapsulates a broad understanding of the world, endowing these models with exceptional versatility and efficiency \cite{jin2024position,bommasani2021opportunities}. 

\begin{table}[!t]
	    \centering
	    \caption{Comparison between our survey and related surveys. (Abbr: Taxonomy means the main taxonomy used in the survey. Standard means standard time series, Spatial means spatial time series, Others include trajectory and event).}
           \setlength\tabcolsep{3 pt}
    	\resizebox{0.8 \linewidth}{!}{
    		\begin{tabular}{cccccc}
    			\toprule
    			Survey & Taxonomy  & Standard  & Spatial & Others\\
    			\midrule
    			  Jin et al. \cite{jin2023large}       & Data & \ding{52} & \ding{52} & \ding{56}\\
    			  Jiang et al. \cite{jiang2024empowering} & Pipeline & \ding{52} & \ding{56} & \ding{52} \\
                    Zhang et al. \cite{zhang2024large} & Pipeline & \ding{52} & \ding{56} & \ding{56}\\
                    Miller et al. \cite{miller2024survey}  & Pipeline & \ding{52} & \ding{56} & \ding{56}\\
    			\midrule
    			\midrule
    			  (Ours)  & \textbf{Methodology} & \textbf{\ding{52}}  & \textbf{\ding{52}} & \textbf{\ding{52}} \\
    			\bottomrule
    		\end{tabular}
    	}
	\label{tab:method_compare}
\end{table}

Inspired by the remarkable achievements of FMs in broad domains like CV and NLP, the concept of Time Series Foundation Models (TSFMs) has garnered attention as a promising direction for time series analysis. TSFMs aim to harness the power of the foundation model paradigm to develop generalized models proficient in understanding and forecasting time series data spanning diverse domains. By capitalizing on large-scale time series datasets, TSFMs hold the promise of attaining superior performance on a spectrum of time series tasks, offering a unified framework that can accelerate research and application developments in this field.

Despite the promising prospects and rapid development of TSFMs, a systematic analysis of TSFMs from a methodological standpoint has been notably absent in prior literature. Existing studies, as depicted in Table~\ref{tab:method_compare}, have concentrated on either the data perspective \cite{jin2023large} or the pipeline perspective \cite{jiang2024empowering} of TSFMs. To bridge this gap, this survey aims to provide a comprehensive methodological analysis of foundation models for learning a variety of time series. This examination will center on scrutinizing their \emph{model architectures}, \emph{pre-training techniques}, \emph{adaptation methods}, and \emph{data modalities}. Through this endeavor, we seek to illuminate an overall picture of core elements in TSFMs, thereby enhancing comprehension regarding the rationale behind their efficacy and the mechanisms driving their substantial potential in time series analysis.

In contrast to previous surveys, this manuscript incorporates the most extensive array of time series data types (see Table \ref{fig:roadmap}), spatial time series, as well as other types such as the trajectory and event. We further summarize the developmental roadmap of current TSFMs in Figure~\ref{fig:roadmap}, in order to foster further innovations and understanding in the dynamic and ever-evolving landscape of TSFMs. In short, our major contributions lie in three aspects:
\begin{itemize}[leftmargin=*]
    \item \textbf{Comprehensive and up-to-date survey.} We offer a comprehensive and up-to-date survey on foundation models for a wide spectrum of time series, encompassing standard time series, spatial time series, and other types (i.e., trajectories and events).
    \item \textbf{Novel methodology-centric taxonomy.} We introduce a novel taxonomy that offers a thorough analysis from a methodological standpoint on TSFMs with the first shot, enabling a full understanding of the mechanism on why and how FMs can achieve admirable performance in time series data.
    \item \textbf{Future research oppotunities.} We discuss and highlight future avenues for enhancing time series analysis using foundation models, urging researchers to delve deeper into this area.
\end{itemize}

\section{Background} 


\textbf{Foundation Models.} 
Foundation models (FMs), also known as large pre-trained models, are a class of deep models that are pre-trained on vast amounts of data, thus equipped with a wide range of general knowledge and patterns. To this end, these models serve as a versatile starting point for various tasks across different domains. Specifically, FMs can be fine-tuned or adapted to specific tasks with relatively small amounts of task-specific data, showcasing remarkable flexibility and efficiency. In CV, FMs such as text-prompted model CLIP \cite{radford2021learning} and visual-prompted model SAM \cite{kirillov2023segment} have propelled advancements in image recognition, object detection, and more. In NLP, FMs such as BERT \cite{DBLP:conf/naacl/DevlinCLT19} and GPT-3 \cite{brown2020language} have revolutionized text understanding and generation tasks. Inspired by the great success of FMs in the above domains, this survey delves into the utilization of these models in the realm of time series analysis.

Concretely, we investigate TSFMs from a methodology perspective: the components of foundation models encompass the data modality, architecture, pre-training, and adaptation technicals: 1) data modality refers to the type of data used for model training, from single modality such as time series, text, images, and audio to multimodality; 2) architecture refers to which deep neural network is adopted as the backbone of FM, with Transformers \cite{vaswani2017transformer,wen2022transformers} being a popular choice for their ability to handle sequential data effectively; 3) Pre-training involves how to train the model on large, diverse datasets to gain a broad understanding of the data, using supervised or self-supervised learning; 4) Adaptation, such as fine-tuning or few-shot learning, is employed to accommodate the pre-trained FMs to specific tasks. This comprehensive framework of FMs, spanning from data modality to adaptation, facilitates the understanding of using them in time series analysis.

\par
\begin{figure}
    \centering
    \includegraphics[width=1 \linewidth]{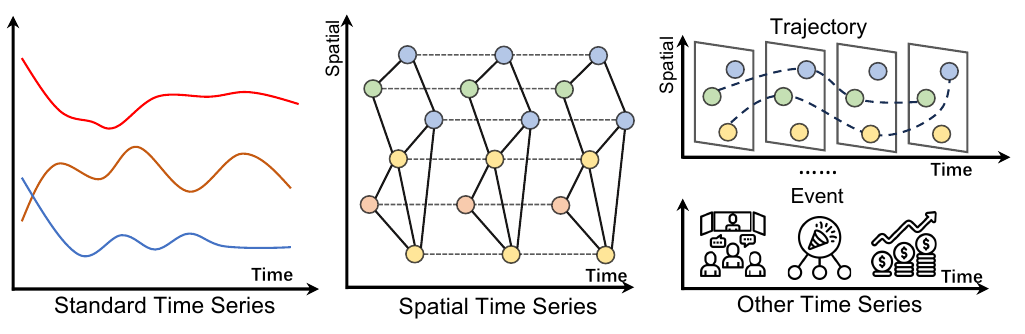}
    \caption{Illustration of various types of time series.}
    \vspace{-0.5em}
    \label{fig:data_type}
    \vspace{-0.5em}
\end{figure}

 \noindent \textbf{Categories of Time Series.}
A time series is commonly described as an ordered sequence of data points. Figure~\ref{fig:data_type} illustrates various types of time series discussed in this survey, including standard time series, spatial time series, trajectories, and events. Note that trajectories and events can be regarded as time series since each data point is associated with a specific timestamp (and location), allowing for analysis using time series techniques such as anomaly detection. These time series are formulated as follows.

%


\par \textit{Definition 1 (Standard Time Series).} The standard time series is defined as a sequence of $T$ data points ordered by time. It can be denoted by $\mathbf{X} = \{\mathbf{x}_1, \mathbf{x}_2, \cdots, \mathbf{x}_T\} \in \mathbb{R}^{T \times D}$, where $\mathbf{x}_t \in \mathbb{R}^{D}$ is the data point at time step $t$, and $D$ is the dimension of each data points. When $D=1$, $\mathbf{X}$ is referred to as a univariate time series, while $D>1$, $\mathbf{X}$ is a multivariate time series.

\textit{Definition 2 (Spatial Time Series).} It refers to a sequence of data points with both temporal and spatial dimensions, which can be represented by $\mathcal{X} = \{\mathbf{X}_1, \mathbf{X}_2, \cdots, \mathbf{X}_T\} \in \mathbb{R}^{N\times T \times D}$, where $\mathbf{X}_t \in \mathbb{R}^{N \times D}$ denotes the signals generated by $N$ sensors with each equipped with $D$ features. Besides, the $N$ sensors are usually associated with spatial correlations, according to which the spatial time series can be further divided into two subtypes: i) spatio-temporal graph, when the spatial correlation of those sensors is described by a graph $G$ with adjacent matrix $\mathbf{A} \in \mathbb{R}^{N \times N}$; ii) spatio-temporal raster, when sensors are distributed uniformly as a grid in geographical space.



\textit{Definition 3 (Trajectory).}   A trajectory is a sequence of timestamped locations that describe the movements of an object in the geographical space. It can be formulated as $\mathcal{T} = \{(l_1, l_2, \cdots, l_T\} \in \mathbb{R}^{T \times 2}$, where $l_t$ means the object's location at time $t$, represented by the two-dimensional coordinates, i.e., latitude and longitude. 


\textit{Definition 4 (Event Sequence).}  An event sequence is a temporally ordered set of events that describe the progression of actions or occurrences within a specific context. It can be formalized as $\mathcal{E} = \{(e_1, t_1), (e_2, t_2), \ldots, (e_n, t_n)\}$, where $e_i$ is an event described by a predicate-argument structure that captures the nature of the occurrence, and $t_i$ denotes the timestamp when $e_i$ occurs.


\input{img/taxonomy}

\section{Taxonomy} 
The proposed taxonomy is illustrated in Figure~\ref{fig:taxonomy}, and the related works can be found in Table~\ref{tab:summary}. The proposed taxonomy unfolds a structured and comprehensive classification to enhance the understanding of foundation models on time series analysis. It is organized into four hierarchical levels, starting with the data category, followed by the model architecture, pre-training techniques, and finally, the application domain. Unlike previous taxonomies, ours distinguishes itself by delving deeper into the foundation models from the methodology perspective, with a keen focus on their architectural designs, pre-training, and adaptation techniques. This method-centric view is pivotal for researchers, providing valuable insights into the mechanisms of why and how foundation models show great potential for time series analysis.

Diving into the details of data categories, we classify the time series data into three distinct types: standard time series, spatial time series, and others, which encompass trajectory and event data. Standard time series data, characterized by their sequential order and temporal dependencies, form the backbone of traditional time series analysis. Spatial time series data, on the other hand, introduce an additional layer of complexity by incorporating geographical or spatial information, making them crucial for applications in urban computing and environmental monitoring. Lastly, the ``others'' category, including trajectory and event data, represents diverse datasets where time plays a critical role, such as the movement of objects over time or the occurrence of specific events, offering a broadened perspective on time series analysis.

From the methodology perspective: i) regarding model architecture, the proposed taxonomy highlights three primary categories: Transformer-based, non-Transformer-based, and diffusion-based models. Transformer-based models leverage self-attention mechanisms to capture long-range dependencies within time series, offering significant advantages in handling sequential data. Non-transformer-based models, with their diverse architectures, cater to a wide range of time series tasks by efficiently processing temporal patterns. Diffusion-based models, a novel addition, employ stochastic processes to model the data generation process, presenting innovative solutions for time series analysis. ii) In terms of pre-training techniques, the proposed taxonomy divides them into fully-supervised and self-supervised methods, the latter of which includes contrastive, generative, and hybrid approaches. This classification shows how different FMs are trained with or without labels. iii) Adaptation strategies, such as zero-shot learning, prompt engineering, tokenization, and fine-tuning, further exemplify the versatility of FMs in customizing to specific time series applications.

\section{Data Perspective}
In this section, we explore advancements in TSFMs from various data perspectives: \emph{standard time series}, \emph{spatial time series}, and \emph{others}. We further categorize our discussion within each subsection into \emph{task-oriented} or \emph{general-purpose} foundation models.

\subsection{Standard Time Series}\label{sec:standard time series}


Standard time series possess diverse properties, including varying sampling rates and temporal patterns, which pose significant challenges in developing relevant FMs. These models aim to identify universal patterns within extensive time series data from varied sources, either to enhance specific tasks or for broad analysis.

Most of the existing attempts are in the category of task-oriented standard time series foundation models. They leverage single or multiple data modalities to craft robust models targeting particular time series tasks, typically forecasting or classification. For models involved only in a single (i.e., time series) modality, they may either be developed from scratch or on existing pre-trained models from other domains like large language or vision models~\cite{zhao2023survey}.

In the first group, Lag-Llama~\cite{rasul2023lag} and TimeGPT-1~\cite{garza2023timegpt} represent pioneering efforts as forecasting foundation models. Both models undergo pre-training on a vast collection of time series data spanning multiple domains. Lag-Llama employs a decoder-only transformer architecture, utilizing lags as covariates, whereas TimeGPT-1 features an encoder-decoder structure with several transformer layers, facilitating efficient zero-shot forecasting. Another noteworthy contribution is TTMs~\cite{ekambaram2024ttms}, a recent endeavor in creating a domain-agnostic forecasting model built upon TSMixer~\cite{ekambaram2023tsmixer}, which itself is pre-trained on diverse time series datasets from various domains. Echoing Lag-Llama's approach, TimesFM~\cite{das2023decoder} emerges as a decoder-only model exhibiting strong zero-shot forecasting capabilities. Concurrently, Moirai~\cite{woo2024unified} introduces an approach with its masked encoder-based universal forecasting transformer, coupled with a new pre-training dataset (LOTSA), containing 27 billion observations from nine distinct domains. Additionally, the exploration extends to diffusion models like TimeGrad~\cite{rasul2021autoregressive} and TransFusion~\cite{sikder2023transfusion}, which primarily focus on optimizing a variational bound on data likelihood, transforming white noise into meaningful samples of the target distribution.

Pre-training from scratch can be expensive, which has spurred the development of alternative approaches that leverage pre-trained models from other domains, such as large language, vision, and acoustic models. For instance, LLM4TS~\cite{chang2023llm4ts} and TEMPO~\cite{cao2023tempo} successfully perform time series forecasting across various datasets by fine-tuning GPT-2~\cite{radford2019language} backbones, predicated on the notion that LLMs can be adapted to process non-linguistic datasets by activating their inherent capabilities. Similarly, Voice2Series~\cite{yang2021voice2series} engages in the synchronization of time series and acoustic data to harness the classification prowess of an acoustic model for time series data. Another approach is presented by Wimmer \textit{et al.}~\cite{wimmer2023leveraging}, who utilize vision-language models (VLMs) to predict market changes. Beyond fine-tuning existing models, a distinct methodology involves direct inference from LLMs for time series forecasting, showcasing commendable zero-shot performance. A notable example of this is LLMTime~\cite{gruver2023large}, which introduces various strategies for effectively tokenizing time series data and transforming discrete token distributions into flexible continuous value densities.

Beyond approaches that focus solely on a single data modality of time series, there have been initiatives towards developing multi-modal, task-oriented foundation models. A notable example is Time-LLM~\cite{jin2023timellm}, which introduces a reprogramming framework to integrate textual and time series information, repurposing an existing LLM into time series forecasters without additional computational costs. In a similar vein, METS~\cite{li2023frozen} employs a trainable ECG encoder alongside a frozen language model to process paired ECG and clinical reports. Further, there is emerging research on directly prompting LLMs for specific time series tasks. For instance, PromptCast~\cite{xue2022promptcast} converts numerical inputs and outputs into prompts, framing the forecasting task as a sentence-to-sentence conversion to leverage language models directly for forecasting. Other studies, such as one involving LLMs prompted with historical stock price data, company metadata, and past economic/financial news, aim to enhance stock return forecasting~\cite{yu2023temporal}. Another example combines GNNs with ChatGPT to predict stock movements~\cite{chen2023chatgpt}, illustrating the diverse applications of these methodologies. Additional noteworthy efforts include \cite{xie2023wall} and \cite{liu2023large}.

Notably, recent efforts have been moved towards creating general-purpose, single-modality standard TSFMs. TS2Vec~\cite{yue2022ts2vec} is a pioneering effort by introducing a universal framework for representing time series via contrastive learning. SimMTM~\cite{dong2023simmtm} explores cross-domain applications, where pre-trained models via masked time series modeling exhibit superior fine-tuning performance in forecasting and classification tasks. More recent works, such as Timer~\cite{liu2024timer} and UniTS~\cite{gao2024units}, further advance the field by facilitating general time series analysis through single, large-scale pre-trained models. 
Moreover, there is a growing interest in adapting pre-trained models, such as LLMs, for broad time series analysis. OFA~\cite{zhou2023one} and TEST~\cite{sun2023test} exemplify this trend, though both approaches necessitate end-to-end fine-tuning for specific tasks.

\subsection{Spatial Time Series}
In complex real-world systems, time series data often display intricate spatial dependencies alongside temporal dynamics, manifesting in forms such as spatio-temporal \emph{graphs} and \emph{rasters}. Similar to the discussion in \shortautoref{sec:standard time series}, research on spatial time series typically encompasses areas such as forecasting and classification. Unlike foundation models for standard time series, most existing research on spatial time series is still in its early stages, often characterized by being domain-specific, single-modality, and task-oriented. In the following, we categorize related works into two specific data modalities and discuss them in different subsections.

\subsubsection{Spatio-Temporal Graph}
Most foundation models for spatio-temporal graphs are task-oriented and only focused on graph data. In the transportation sector, TFM~\cite{wang2023building} utilizes graph structures and algorithms to analyze the behavior and interactions within transportation systems, showing promising results in urban traffic forecasting. ST-LLM~\cite{liu2024spatial} combines spatio-temporal information with a partially frozen LLM to improve traffic predictions, while DiffSTG~\cite{wen2023diffstg} applies denoising diffusion models to spatio-temporal graphs for probabilistic traffic forecasting. Efforts towards domain-agnostic models include STEP~\cite{shao2022pre}, which links spatio-temporal GNNs with a pre-trained transformer for enhanced forecasting by learning from extensive historical data. Similarly, STGCL~\cite{liu2022contrastive} and SPGCL~\cite{li2022mining} explore the integration of contrastive learning into spatio-temporal graph forecasting, indicating its potential benefits. Research on general-purpose models for spatio-temporal graphs is limited. A notable example, USTD~\cite{hu2023unifying}, introduces a unified model for both forecasting and kriging tasks, employing an uncertainty-aware diffusion approach to address diverse challenges effectively.

\subsubsection{Spatio-Temporal Raster}
Spatio-temporal raster refers to a data modality that captures and organizes spatial information over various time points in a grid-like format. This modality is primarily utilized in climate foundation models. For instance, FourCastNet~\cite{pathak2022fourcastnet} is a global, data-driven weather forecasting model delivering accurate short to medium-range predictions worldwide. Similar models, such as FengWu~\cite{chen2023fengwu} and W-MAE~\cite{man2023w}, follow suit. Notably, Pangu-Weather~\cite{bi2022pangu}, which is trained on 39 years of global data, achieves superior deterministic forecasting outcomes across all evaluated variables compared to leading numerical weather prediction systems. On a different note, ClimaX~\cite{nguyen2023climax} aims at general-purpose climate foundation models, pre-trained with diverse datasets covering various variables, spatio-temporal scopes, and physical contexts. It is designed for fine-tuning across a wide array of climate and weather-related tasks, such as forecasting, projection, and downscaling, even for atmospheric variables and spatio-temporal scales not encountered during its pre-training phase. However, there is a scarce number of domain-agnostic models for spatio-temporal raster data. DYffusion~\cite{cachay2023dyffusion}, for example, capitalizes on the temporal dynamics inherent in raster data, integrating these dynamics directly with the model's diffusion steps to create a stochastic, time-conditioned interpolator and forecasting network.

\subsection{Others}
In addition to standard and spatial time series, various other types of data incorporate the temporal dimension, including trajectories, events, and clinical records. A majority of studies in this category focus on trajectory data. For Transformer-based models, AuxMobLCast \cite{xue2022leveraging} fine-tunes pre-trained LLMs through mobility prompting and auxiliary  POI Category classification to forecast human mobility patterns, effectively bridging the gap between natural language processing and temporal sequence prediction. LLM-Mob \cite{wang2023i}  encodes human mobility data into structured prompts that instruct LLMs to consider both long-term and short-term behavioral patterns, along with time-specific context, to generate accurate and explainable predictions of future locations. For non-Transformer-based models, Trembr \cite{fu2020trembr} leverages auto-encoding techniques to extract road network and temporal information embedded in trajectories effectively. While START~\cite{jiang2022self} introduces a hybrid approach to trajectory embedding learning by combining masked language model~\cite{DBLP:conf/naacl/DevlinCLT19} and SimCLR~\cite{DBLP:conf/icml/ChenK0H20} to enhance its learning capability. More recently, GTM \cite{lin2024gtm} separates trajectory features into three domains, which can be masked and generated independently to meet specific input and output requirements of a given task. Then, GTM is pre-trained by reconstructing densely sampled trajectories in an auto-regressive manner given re-sampled sparse counterparts.  For the diffusion-based model, DiffTraj \cite{zhu2024difftraj} reconstructs and synthesizes geographic trajectories from white noise through a conditioned reverse trajectory denoising process.

\section{Methodology Perspective} 

In this section, we dissect TSFMs from a methodology perspective, focusing on \emph{architecture} and \emph{pipeline} (including pre-train and adaptation) intricacies. This discussion aims to elucidate the intricate mechanisms driving these models' efficacy and adaptability.

\begin{figure}[t]
    \centering
    \includegraphics[width=\linewidth]{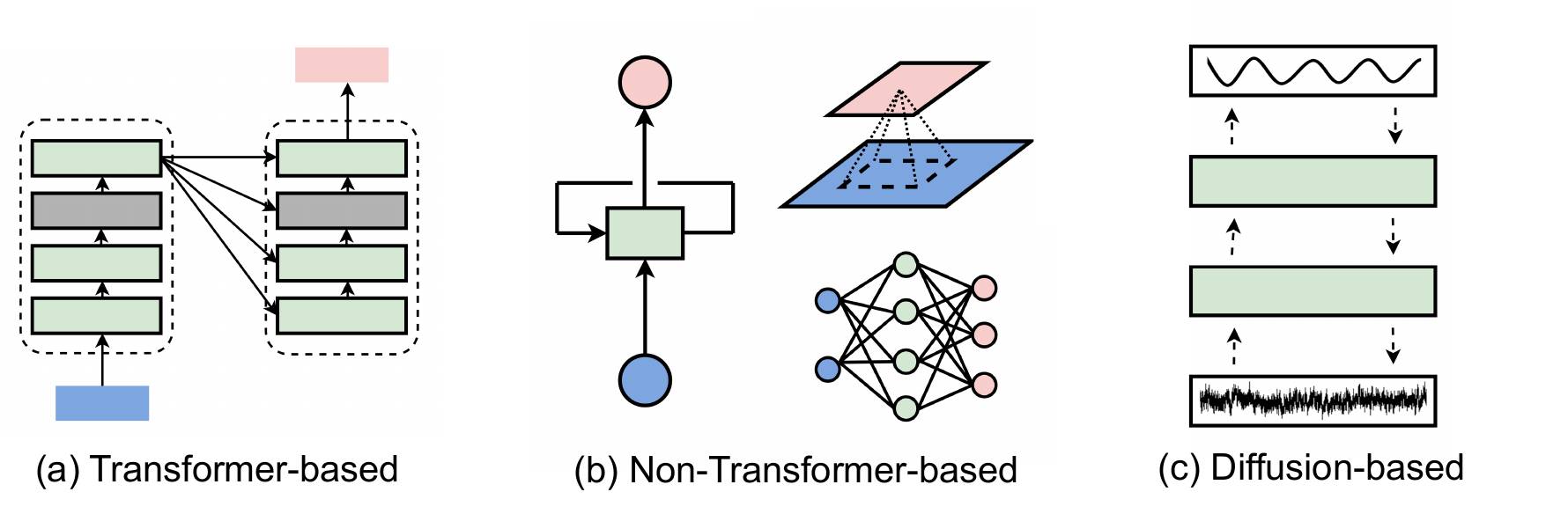}
    \caption{Architectures of TSFMs.}
    \label{fig:arch}
\end{figure}

\subsection{Architecture}
As shown in Figure~\ref{fig:arch}, we first delve into the architecture of TSFMs, including \emph{Transformer-based models}, \emph{non-Transformer-based nodels} and \emph{diffusion-based models}, focusing on the underlying mechanisms that shape their capabilities, as well as how they could be applied on various time series.

\subsubsection{Transformer-based Models}

The architecture of FMs has seen a significant convergence towards the Transformer \cite{vaswani2017transformer}, a model architecture first introduced for NLP tasks.
The core innovation of the Transformer lies in its utilization of the attention mechanism, which allows the model to dynamically focus on different parts of the input data. The attention function can be succinctly described as
$\texttt{Attention}(Q,K,V) = \texttt{Softmax}(Q K^{T}/\sqrt{d_k}) V$, 
where $Q$, $K$, and $V$ represent the queries, keys, and values matrices respectively, each with dimensions $T\times d_k$, and $d_k$ serves as a scaling factor to moderate the dot products' magnitude. It is evident from the formula that the attention mechanism has the capability to learn global, long-range dependencies in data. This distinguishes it from previous architectures, which were often limited by their local receptive fields or dependency windows. Besides, the Transformer's design is inherently friendly to parallelization, which allows for significant scalability, enabling the processing of large datasets and the construction of models with billions of parameters. Such scalability and efficiency in capturing intricate data patterns have led to the widespread adoption of the Transformer architecture beyond its initial application in natural language processing (NLP) \cite{DBLP:conf/naacl/DevlinCLT19} to fields including computer vision (CV), speech, video, time series (Table \ref{tab:summary})  and beyond. 

The choice of foundation model framework remains debated in the realm of time series analysis, contrasting the trend towards decoder-only models in natural language processing. Notable works in this area includes encoder-only \cite{nie2022time, woo2024unified, gao2024units}, encoder-decoder \cite{dong2024timesiam, garza2023timegpt, ansari2024chronos}, and decoder-only \cite{das2023decoder, rasul2023lag, liu2024timer} models. \textit{Ansari et al.} \cite{ansari2024chronos} analyze the applicability of the encoder-decoder framework to decoder-only models. \textit{Liu et al.} \cite{liu2024timer} discuss that while the encoder-only model is favored in time series forecasting for its effectiveness on small datasets, the decoder-only architecture, with its strong generalization and capacity, could be preferred for large-scale time series models. The diversity in the architectural choices underscores the potential and necessity for further exploration within this field.

In terms of standard time series analysis, the Transformer architecture leverages its sequence modeling capabilities to capture temporal dynamics. This includes either repurposing pretrained LLMs for time series to leverage their preexisting sequence modeling strengths \cite{xue2022promptcast}, or directly using Transformer as a base for TSFMs, training from scratch to achieve models best suited for the specifics of time series data \cite{garza2023timegpt}. Besides, various techniques have been innovated to augment the functionality of Transformer models in time series analysis comprehensively. A common practice in TSFMs segments time series into patches, which can effectively encapsulate local dynamics within input tokens \cite{zhou2023one, chang2023llm4ts, das2023decoder, jin2023timellm, rasul2023lag, nie2022time, cao2023tempo, woo2024unified, cheng2024nuwats}. Another critical design is the normalization layer, where reversible instance normalization \cite{kim2021reversible} techniques, standardizing data through instance-specific mean and variance then reverting it at the output layer, have found extensive application across the above models. Moreover, specialized approaches such as multi-resolution analysis, exemplified by Moirai \cite{woo2024unified} through the employment of varying patch sizes, and decomposition strategies, as implemented by TEMPO \cite{cao2023tempo} via the separation of complex interactions into the trend, seasonal, and residual components, have been shown to enhance model efficacy substantially.




For spatial time series, the attention mechanism is utilized to model both the spatial and temporal dependency. For instance, ST-LLM \cite{liu2024spatial} employs a novel partially frozen attention strategy for traffic prediction, leveraging spatial-temporal embeddings to capture the intricate dynamics of traffic data across space and time. Conversely, other studies opt for independent modeling of spatial and temporal relationships. TFM \cite{wang2023building} is a case in point, which employs attention mechanisms within a dynamic graph encoder for spatial modeling, integrating time encoding for temporal aspects, embodying principles of transformers in addressing traffic system's spatial-temporal dependencies. Besides simultaneously modeling spatial and temporal relationships, there exists an alternative approach that augments the Transformer model with additional spatial models or external spatial information to enhance its capabilities in the temporal modeling of time series. An example of this is STEP \cite{shao2022pre}, which uses unsupervised pre-trained TransFormer blocks to model temporal relationship from long-term history time series, while applying a graph structure learner and spatio-temporal GNNs based on the representation of TransFormer blocks. Furthermore, the application of Transformer models extends to the domain of spatial-temporal prompt learning, as evidenced by initiatives such as MetePFL \cite{chen2023prompt} and FedWing \cite{chen2023spatial}.



In addition to conventional time series data, the Transformer architecture has demonstrated efficacy across a diverse array of temporal datasets, such as trajectory and healthcare records, as summarized in Figure \ref{fig:taxonomy}. This expansion highlights the Transformer's versatile capacity for temporal data analysis.

\subsubsection{Non-Transformer-based Models}

Excluding the widespread adoption of Transformers, a diverse array of traditional pre-training methods leveraged models such as Multi-Layer Perceptrons (MLPs) \cite{ekambaram2024ttms}, Recurrent Neural Networks (RNNs) \cite{fu2020trembr}, and Convolutional Neural Networks (CNNs) \cite{wu2022timesnet} as the backbone for pre-training. These models, each with their unique strengths, are notable for their effectiveness in both conventional and spatial time series data.

MLPs and CNNs are both acclaimed for their capabilities in modeling spatial and temporal data effectively. CNN-based architectures, in particular, have garnered significant attention in self-supervised learning for general time series representation, with a notable emphasis on the usage of ResNet \cite{zhang2022self, dong2023simmtm} and dilated convolution layers \cite{yue2022ts2vec, ozyurt2022contrastive} as foundational backbones. Those approaches predominantly employ 1D convolutional operations. In contrast, TimesNet \cite{wu2022timesnet} introduces a novel perspective by converting 1D time series data into 2D tensors, facilitating the adaptive identification of multi-periodicity and the extraction of complex temporal variations through the use of a parameter-efficient inception block. MLP-based models, on the other hand, are lauded for their lightweight design, offering benefits in terms of reduced computational time and cost. TSMixer \cite{ekambaram2023tsmixer} and TTMs \cite{ekambaram2024ttms}, as instances, both claiming superior efficiency in memory usage and processing speed while still delivering competitive performance.


RNNs have been acknowledged for their proficiency in temporal data modeling \cite{fu2020trembr,hewamalage2021recurrent}. Recently, there has been a resurgence of interest in RNN architectures, which poses a compelling challenge to the prevailing Transformer-based models. This trend is driven by the quest for models that are not only more resource-efficient but also adept at handling longer sequences through their inherent linear complexity. A notable embodiment is the RWKV-TS \cite{hou2024rwkv}, which leverages the RWKV \cite{peng2023rwkv}, an RNN-type foundation model architecture, demonstrating promising potential for general time series analysis. This emerging trend presents a valuable opportunity for time series research and applications.





\subsubsection{Diffusion-based Models}


Diffusion-based foundation models have gained prominence in CV \cite{rombach2022high, peebles2023scalable} and video \cite{videoworldsimulators2024} due to their proficiency in learning complex data distributions, yet their exploration in time series analysis remains nascent. These models function by gradually introducing and then reversing noise to data, effectively learning the generative process of original data through the reverse diffusion process. This unique mechanism equips diffusion models with great potential to serve as versatile foundation models capable of tackling prediction, imputation, and anomaly detection in time series. 

In standard time series and other temporal data, diffusion models predict future states by capturing temporal dynamics, generating smooth transitions from current to potential future states \cite{rasul2021autoregressive,wang2023diffload}. Applied to spatial time series, they extend this capability to model spatial correlations alongside temporal ones, providing insights into the interplay between space and time, particularly beneficial in fields like traffic forecasting \cite{wen2023diffstg}.

\subsection{Pipeline}
In this part, we review TSFMs from the pipeline perspective, including diverse model acquisition and adaptation mechanisms.

\subsubsection{Pre-training}
Pre-training is an initial and crucial step for building TSFMs, since the knowledge learned in this phase enables the models to generalize across different contexts and quickly adapt to various downstream tasks with minimal adaptations. On the other hand, the diverse nature of pre-training data (\textit{e.g.}, standard time series, spatial time series, and trajectories), as well as the way the data is used, lead to a wide spectrum of pre-training mechanisms when building and deploying foundation model. In this survey, we propose a new perspective mostly based on learning objectives in the pre-training phase, to categorize existing methods for TSFMs. These mechanisms include fully-supervised, self-supervised (generative, contrastive, hybrid of generative and contrastive), and others.

Fully-supervised pre-training refers to the strategy where the foundation model is initially trained on one or multiple large time series datasets with labels to capture the complex temporal dynamics and learn generalizable representations. TTMs~\cite{ekambaram2024ttms} proposes a universal time series foundation model supervised framework that is able to handle the heterogeneity of multiple time series datasets and effectively build the forecasting capability during pre-training, via the design of multi-resolution enhancements (\textit{e.g.}, adaptive patching, data augmentation via downsampling, \textit{etc.})
Fully-supervised pre-training for TSFMs is particularly suited for scenarios where there is sufficient labeled historical data. Moreover, this pre-training technique is more frequently used in some domain-specific applications such as transportation~\cite{duan2019pre, wang2023building} and climate~\cite{pathak2022fourcastnet, bi2023accurate}, where the model can be directly tailored for downstream forecasting tasks with the ease of minimal adaptations.

We categorize the generative pre-training strategy as a general modeling of time series representations, including reconstruction and probabilistic modeling of time series inputs. In reconstruction-based pre-training, an effective learning objective is to recover the original input space via masked autoencoding strategies~\cite{shao2022pre,chen2023prompt}. In the probabilistic modeling methods, the latent representation space formed from temporal or spatial-temporal encoders is optimized toward an estimated density via maximizing log-likelihood, based on which the forecasts can be sampled~\cite{rasul2021autoregressive,wen2023diffstg}. Moreover, it is also beneficial to leverage contrastive learning to enhance the robustness of pre-training time series foundation models. The key is to construct and utilize the self-supervision signals by generating informative positive pairs as well as filtering out unsuitable negative pairs when performing augmentation~\cite{liu2022contrastive}. 
In addition to the aforementioned two self-supervised strategies, the efficacy of the hybrid variant has also been validated, where the pre-trained model on fewer time series data outperforms the supervised counterpart~\cite{jiang2022self}.

In general, self-supervised pre-training enables foundation models to exploit the vast amounts of unlabeled time series data, providing generic temporal knowledge that can be further fine-tuned for specific downstream tasks. Compared with fully-supervised pre-training, it provides a more generic and realistic solution for the acquisition of a time series foundation model. 

Note that the above pre-training methods typically build the model from scratch and
obtain the universal knowledge from data with the same modality (\textit{i.e.}, time series). Nevertheless, recent advancements in time series research have heightened the usage of LLMs~\cite{zhou2023one, jin2023timellm, xue2022promptcast, chang2023llm4ts, cao2023tempo, gruver2023large, yu2023temporal, chen2023chatgpt, xie2023wall, liu2023large, liu2024autotimes, liu2024spatial, xue2022leveraging, wang2023i, shi2023language, gunjal2023drafting}, VLMs~\cite{wimmer2023leveraging}, and AMs~\cite{yang2021voice2series} that are pre-trained from other data modalities (text sequence, image-text sequence, acoustic signals).

\subsubsection{Adaptation}
\begin{figure}[!h]
    \centering
    \includegraphics[width=1 \linewidth]{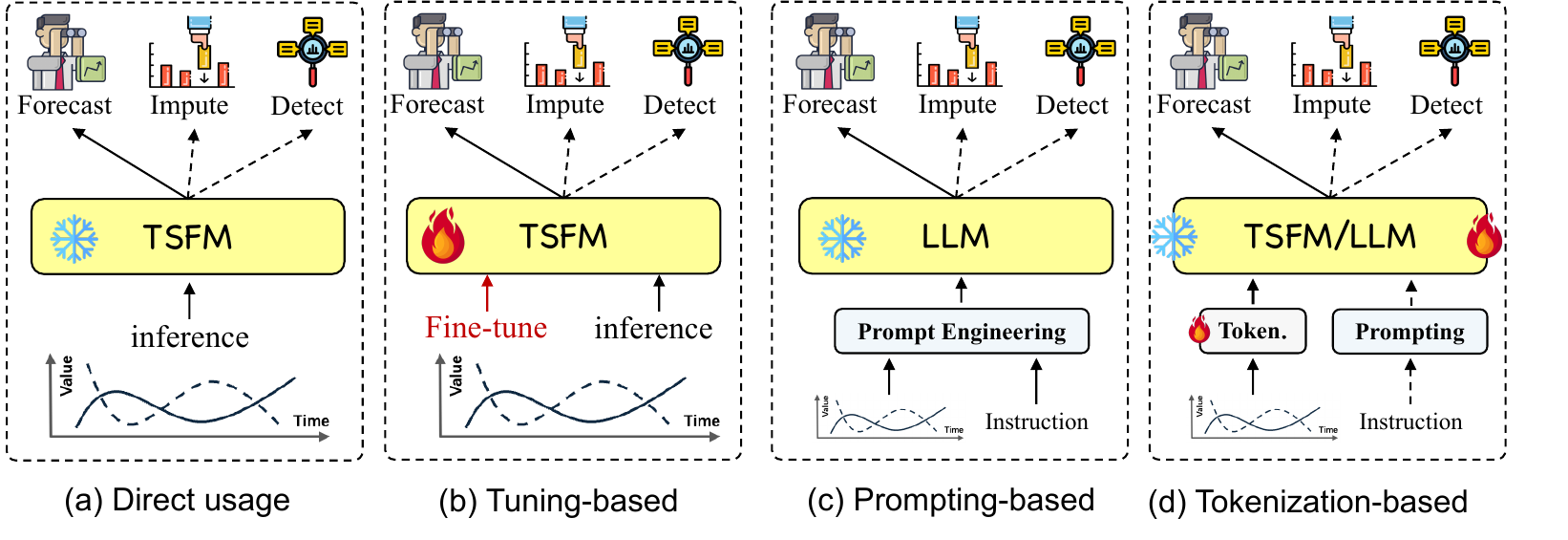}
    \caption{Illustration of different adaptation techniques.}
    \label{fig:adaptation}
\end{figure}

The adaptation phase tailors TSFM to specific tasks or datasets, enhancing its performance on those tasks by leveraging the learned generic temporal knowledge. We partition prior methods into four branches, including direct usage, fine-tuning, prompt engineering, and time series tokenization (see Figure \ref{fig:adaptation}).

Direct usage (also called zero-shot), means no further fine-tuning on the target datasets, suggesting the sufficient capability of a pre-trained model for downstream tasks. It can also indicate the homogeneity between the pre-trained dataset and target dataset, especially for some real-world applications where a foundation model is built to fulfill domain-specific tasks~\cite{bi2022pangu,chen2023fengwu}.

Fine-tuning is a common strategy to adapt foundation models to target tasks. Based on the way the foundation model is used on the target dataset, there are three mainstream works: fine-tuning the whole model~\cite{man2023w,nguyen2023climax,pathak2022fourcastnet} or specific components (\textit{e.g.}, training positional embeddings and layer normalization, while keeping feedforward and attention layers frozen when fine-tuning LLMs)~\cite{chang2023llm4ts,zhou2023one}, to directly infer results, or integrate foundation models as part of the whole model~\cite{sun2023test,chen2023chatgpt,xue2022leveraging,li2023frozen}.

Prompt engineering is more specialized in LLM-based TSFMs. The prompt can be handcrafted with task-specific textual input and directly used to query the output for downstream prediction~\cite{xue2022promptcast,wang2023i} or intermediate embedding as feature enhancement~\cite{xue2022leveraging}. Besides, the prompt can also be parameterized vectors and end-to-end learnable when optimizing the model on target datasets~\cite{sun2023test,cao2023tempo}. In comparison to static prompts, the use of trainable prompts enhances the ability of LLMs to comprehend and match the context of given time series inputs. For example, TEMPO~\cite{cao2023tempo} constructs a trainable prompt pool with distinct key-value pairs, and retrieves the most representative prompt candidates with the highest similarity scores. 

Time series tokenization aims to effectively represent the time series as embeddings, which is also more frequently adopted in transformer-based architectures~\cite{jin2023timellm,cao2023tempo,nie2022time}. Common tokenization techniques include reversible instance normalization~\cite{kim2021reversible} that mitigates distribution shift, patching with channel independence strategy that effectively and efficiently extracts the time series context~\cite{nie2022time},  as well as the joint usage of time series decomposition to explicitly represent explainable components~\cite{cao2023tempo} for the ease of subsequent temporal modeling. 

In addition to the main branches of adapting TSFMs, it is also worth noting that some fine-tuning strategies take real-world constraints into account. For example, the fine-tuning is performed in a privacy-preserved manner~\cite{chen2023prompt,chen2023spatialtemporal}.

\subsection{Modality}
During the pre-training/adaptation of TSFMs, prior methods involve single or multiple data modalities, where standard time series, trajectory, raster, and text can be treated as different forms with unique domain perspectives. In this part, we review the data modalities that are used in existing TSFMs across different domains.

\subsubsection{Single-modality}
A majority of current TSFMs are constructed and tailored on the basis of single-modal data. Compared with multi-modal methods, the single-modal time series modeling strategy gains the advantages of inherent simplicity and bypasses the challenges of handling modality gaps, yet frequently demonstrates excellent empirical results across a wide range of real-world applications, such as traffic~\cite{shao2022pre,liu2024spatial} and climate forecasting~\cite{chen2023fengwu,nguyen2023climax}.

\subsubsection{Multi-modality}
However, the single-modal methods may not encapsulate the full picture for several challenging downstream tasks in finance~\cite{chen2023chatgpt,yu2023temporal} and healthcare domains~\cite{liu2023large,li2023frozen}. To cope with this issue, there have been initiatives towards developing multi-modal, task-oriented FMs, where additional information provides useful information to enhance the model capability. In \textit{Chen et al.}, an external ChatGPT is queried to construct an evolving graph structure representing companies, based on the analysis of news headlines at specific time steps. As such, the inferred graph and stock prices are fed into the time series model (that uses GNN and LSTM for information propagation) to generate stock price movement predictions. Another example in healthcare also demonstrated the effectiveness of multi-modal medical context modeling, which aligns the embedding of ECG (Electrocardiogram) and corresponding medical text reports under a self-supervised contrastive learning framework and performs ECG classification. In general multi-modal time series analysis, similar cross-modality alignment strategies (\textit{e.g.,} contrastive learning~\cite{sun2023test}, reprogramming~\cite{jin2023timellm}, token-wise prompting~\cite{liu2024autotimes}) are adopted, where the multi-modal inputs are often the textual description of datasets and pre-training word embedding from LLMs. As a notable example, Time-LLM~\cite{jin2023timellm} introduces a reprogramming framework that aligns the language knowledge from pre-trained word embedding and time series information via linear projection and multi-head attention, where the handcrafted dataset descriptions are also used to quired text token embeddings as prompts, which further enhances the embedding space and informs the LLM to comprehend the task contexts. As such, utilizing multi-modal data facilitates the repurpose of the existing LLM into time series forecasters without additional computational costs.

\section{Conclusion}
The rapid development of FMs has revolutionized the research fields in different domains. In this survey, we provide a comprehensive and updated review of FMs specifically designed for time series analysis.  A novel taxonomy is proposed from a methodology-centric perspective by classifying FMs based on key components including model architecture, pre-training technique, adaptation technique, and data modality.  Our survey facilitates understanding the underlying mechanism of applying the FMs to time series. Furthermore, we believe that the consolidation of the latest advancements, as well as the potential future direction (see Appendix), can inspire more innovative works within the field of time series analysis.

\section*{Acknowledgements}
This work is mainly supported by the Guangzhou-HKUST(GZ) Joint Funding Program (No. 2024A03J0620). It is also funded by Guangzhou Municipal Science and Technology Project 2023A03J0011.
\vspace{-1em}
\bibliographystyle{ACM-Reference-Format}
\bibliography{ref}

\vspace{-0.5em}
\appendix

\section{Appendix}
In this section, we discuss the future research directions and opportunities of TSFMs from the methodology perspective.

\textbf{Incooporating Multi-modalities.} As illustrated in this survey, a majority of current foundation models for time series are developed based on a single modality. However, many real-world dynamic systems are coupled with various modalities (time series, text, even image data). It would be a promising direction to leverage various modalities along with the time series in TSFM to learn more comprehensive and generalized knowledge, therefore significantly boosting the performance of different downstream tasks.

\textbf{Exploring more Efficient Architectures.} Currently, the Transformer serves as the dominant architecture for building the foundation model. Though promising,  Transformer-based foundation models have quadratic scaling with respect to the sequence length due to their self-attention mechanism. This makes them computationally expensive and memory-intensive for processing long sequences. Therefore, it is an interesting avenue for future study to explore more efficient FM backbone architectures, such as state-space models Mamba \cite{gu2023mamba}.

\textbf{Developing more Effective Pipelines.} Time series data has unique properties such as temporal distribution shift \cite{adarnn,zhou2023maintain,wang2023st} (i.e., the data distribution will evolve over time) and causality (i.e., casual relationship can exist between different points in the time series) \cite{xia2024deciphering}. Therefore, it would be another existing as well as challenging future direction to develop TSFMs that can well address the temporal distribution shift or have a powerful Interpretability for downstream tasks.

\textbf{Protecting Privacy.} Protecting privacy is an essential concern when training foundation models on diverse sources and modalities of data, which raises potential risks of exposing sensitive information. As such, one future direction is the development of robust privacy-preserving techniques for training the TSFM from multi-source datasets, as well as keeping the utility of the trained FMs.  This may include the advancement of federated learning approaches, where models can be trained across multiple decentralized devices or servers without exchanging raw data.


\begin{table*}[h]
    \centering
    \footnotesize
    \caption{Summary of representative foundation models tailored for time series data modeling. N/A means not not applicable} 
    \resizebox{\textwidth}{!}{
    \begin{tabular}{c|cccccccccc}
    \toprule\toprule	
    \textbf{} & \textbf{Category} & \textbf{Method} & \textbf{Modaility} & \textbf{Pre-training} & \textbf{Adaptation} & \textbf{Domain} & \textbf{Year} \\
    \hline
    \multirow{30}{*}{\rotatebox[origin=c]{90}{\textbf{Standard Time Series}}}
    & \multirow{24}{*}{\rotatebox[origin=c]{0}{Transformer-based}} 
    & Time-LLM~\cite{jin2023timellm} & Multi-Modality & Pretrained LLM & Prompt-engineering \& Tokenization & General & 2023 \\
    & & OFA~\cite{zhou2023one} & Single-Modality & Pretrained LLM & Fine-tuning & General & 2023 \\
    & & PromptCast~\cite{xue2022promptcast} & Multi-Modality & Pretrained LLM & Prompt-engineering & General & 2022 \\
    & & TEST~\cite{sun2023test} & Multi-Modality & Contrastive & Prompt-engineering \& Tokenization & General & 2023 \\
    & & LLM4TS~\cite{chang2023llm4ts} & Single-Modality & Pretrained LLM & Fine-tuning & General & 2023 \\
    & & TEMPO~\cite{cao2023tempo} & Single-Modality & Pretrained LLM & Fine-tuning \& Prompt-engineering & General & 2023 \\
    & & LLMTime~\cite{gruver2023large} & Single-Modality & Pretrained LLM & Zero-shot & General & 2023 \\
    
    & & Yu \textit{et al.}~\cite{yu2023temporal} & Multi-Modality & Pretrained LLM & Zero-shot & Finance & 2023 \\
    & & Chen \textit{et al.}~\cite{chen2023chatgpt} & Multi-Modality & Pretrained LLM & Zero-shot & Finance & 2023 \\
    & & Xie \textit{et al.}~\cite{xie2023wall} & Multi-Modality & Pretrained LLM & Zero-shot & Finance & 2023 \\
    & & Wimmer \textit{et al.}~\cite{wimmer2023leveraging} & Single-Modality & Pretrained VLM & Fine-tuning & Finance & 2023 \\
    & & Liu \textit{et al.}~\cite{liu2023large} & Multi-Modality & Pretrained LLM & Prompt-engineering & Healthcare & 2023 \\
    & & METS~\cite{li2023frozen} & Multi-Modality & Contrastive  & Zero-shot & Healthcare & 2023 \\
    & & Voice2Series~\cite{yang2021voice2series} & Single-Modality & Pretrained AM  & Tokenization & General & 2021 \\
    & & PatchTST~\cite{nie2022time} & Single-Modality & Generative & Fine-tuning & General & 2022 \\
    & & Moirai~\cite{woo2024unified} & Single-Modality & Generative & Fine-tuning & General & 2024 \\
    & & Timer~\cite{liu2024timer} & Single-Modality & Generative & Fine-tuning & General & 2024 \\
    & & TimeSiam~\cite{dong2024timesiam} & Single-Modality & Generative & Fine-tuning & General & 2024 \\
    & & TimeXer~\cite{wang2024timexer} & Single-Modality & Fully-supervised & - & General & 2024 \\
    & & AutoTimes~\cite{liu2024autotimes} & Multi-Modality & Pretrained LLM & Prompt-engineering \& Tokenization & General & 2024 \\
    & & Lag-Llama~\cite{rasul2023lag} & Single-Modality & Generative & Zero-shot \& Fine-tuning & General & 2023 \\
    & & Das \textit{et al.}~\cite{das2023decoder} & Single-Modality & Generative & Zero-shot & General & 2023 \\ 
    & & TimeGPT-1~\cite{garza2023timegpt} & Single-Modality & Generative & Zero-shot \& Fine-tuning & General & 2023 \\
    & & UniTS~\cite{gao2024units} & Single-Modality & Fully-supervised \& Generative & Zero-shot \& Prompt-engineering & General & 2024 \\
    & & Chronos~\cite{ansari2024chronos} & Single-Modality & Generative & Zero-shot & General & 2024 \\
    & & SimMTM~\cite{dong2023simmtm} & Single-Modality & Hybrid
    & Fine-tuning & General & 2023 \\
    & & MTSMAE~\cite{tang2022mtsmae} & Single-Modality & Generative & Fine-tuning & General & 2022 \\
    & & TimeCLR~\cite{yeh2023toward}  & Single-Modality & Contrastive & Fine-tuning & General & 2023 \\
     & & UniTime~\cite{liu2024unitime}  & Single-Modality & Pretrained LLM & Fine-tuning & General & 2024 \\
     & & GTT~\cite{feng2024curve}  & Single-Modality & Fully-supervised & Zero-shot & General & 2024 \\
    
   
    \cline{2-8}
    & \multirow{7}{*}{\rotatebox[origin=c]{0}{Non-Transformer-based}} 
    & TF-C~\cite{zhang2022self} & Single-Modality & Contrastive  & Fine-tuning & General & 2022 \\
    & & CLUDA~\cite{ozyurt2022contrastive} & Single-Modality & Contrastive & Fine-tuning & General & 2022 \\
    & & TS2Vec~\cite{yue2022ts2vec} & Single-Modality & Contrastive  & Fine-tuning & General & 2021 \\
    & & TimesNet~\cite{wu2022timesnet} & Single-Modality & Fully-supervised & - & General & 2022 \\
    & & TSMixer~\cite{ekambaram2023tsmixer} & Single-Modality & Generative & Fine-tuning & General & 2023 \\
    & & TTMs~\cite{ekambaram2024ttms} & Single-Modality & Fully-supervised & Zero-shot \& Fine-tuning \& Prompt-engineering & General & 2024 \\
    & & RWKV-TS~\cite{hou2024rwkv} & Single-Modality & Fully-supervised & - & General & 2024 \\

     \cline{2-8}
    & \multirow{10}{*}{\rotatebox[origin=c]{0}{Diffsuion-based}}
    &   {TimeGrad}~\cite{rasul2021autoregressive}  & Single-Modality & Generative & N/A & General & 2021 \\
    & & {D$^3$VAE}~\cite{li2022generative} & Single-Modality & Generative & N/A & General & 2022 \\
    & & {TransFusion}~\cite{sikder2023transfusion} & Single-Modality & Generative & N/A & General & 2023 \\
    & & {ScoreGrad}~\cite{yan2021scoregrad}  & Single-Modality & Generative & N/A & General & 2021 \\
    & & Biloš et al.~\cite{bilovs2022modeling} & Single-Modality & Generative & N/A & General & 2022 \\
    & & Crabbé et al.~\cite{crabbe2024time} & Single-Modality & Generative & N/A & General & 2024 \\
    & & TimeDiff~\cite{shen2023non}  & Single-Modality & Generative & N/A & General & 2023 \\
    & & Wang et al. ~\cite{wang2023observed} & Single-Modality & Generative & N/A & General & 2023 \\
    & & DiffTime~\cite{coletta2024constrained} & Single-Modality & Generative & N/A & General & 2024 \\
    & & {FTS-Diffusion}~\cite{huang2024generative} & Single-Modality  & Generative & N/A & Finance & 2023 \\
    & & {DiffLoad}~\cite{wang2023diffload} & Single-Modality  & Generative & N/A & Power & 2023 \\

    \hline
    \multirow{24}{*}{\rotatebox[origin=c]{90}{\textbf{Spatial Time Series}}}
    & \multirow{14}{*}{\rotatebox[origin=c]{0}{Transformer-based}}
    & CPPBTR~\cite{duan2019pre} & Single-Modality & Fully-supervised & Fine-tuning & Transportation & 2019 \\
    & & TFM~\cite{wang2023building} (graph TF) & Single-Modality & Fully-supervised & - & Transportation & 2023 \\
    & & STEP~\cite{shao2022pre} & Single-Modality & Generative & Fine-tuning & Transportation & 2022 \\
    & & ST-LLM~\cite{liu2024spatial} & Single-Modality & Pretrained LLM & Fine-tuning & Transportation & 2024 \\
    & & FourCastNet~\cite{pathak2022fourcastnet} & Single-Modality & Fully-supervised & Fine-tuning & Climate & 2022 \\
    & & MetePFL~\cite{chen2023prompt} & Single-Modality & Generative & Federated Learning \& Prompt-Learning & Climate & 2023 \\
    & & FedWing~\cite{chen2023spatial} & Single-Modality & Fully-supervised & Federated Learning \& Prompt-Learning & Climate & 2023 \\
    & & ClimaX~\cite{nguyen2023climax} & Single-Modality & Fully-supervised & Fine-tuning & Climate & 2023 \\
    & & FengWu~\cite{chen2023fengwu} & Single-Modality & Generative & Zero-shot  & Climate & 2023 \\
    & & Pangu-Weather~\cite{bi2023accurate} & Single-Modality & Fully-supervised & Zero-shot & Climate & 2023 \\
    & & W-MAE~\cite{man2023w} & Single-Modality & Generative & Fine-tuning & Climate & 2023 \\
    & & GATGPT~\cite{chen2023gatgpt} & Single-Modality & Pretrained LLM & Fine-tuning & General & 2023 \\
    & & TPLLM~\cite{ren2024tpllm} & Single-Modality & Pretrained LLM & Fine-tuning & Transportation & 2024 \\
     & & UniST~\cite{yuan2024unist} & Single-Modality & Generative & Fine-tuning & General & 2024 \\
    \cline{2-8}
    
    & \multirow{2}{*}{\rotatebox[origin=c]{0}{Non-Tranformer-based}}
    & SPGCL~\cite{li2022mining} & Single-Modality & Contrastive & - & General & 2022 \\ 
    & & STGCL~\cite{liu2022contrastive} & Single-Modality & Contrastive & Fine-tuning & General & 2021 \\

    \cline{2-8}
    & \multirow{6}{*}{\rotatebox[origin=c]{0}{Diffusion-based}}
    &  DiffSTG~\cite{wen2023diffstg} & Single-Modality & Generative & N/A & General & 2023 \\
    & & DSTPP~\cite{yuan2023spatio} & Single-Modality & Generative & N/A & General & 2023 \\
    & & DYffusion~\cite{cachay2023dyffusion} & Single-Modality & Generative & N/A & General & 2023 \\
    & & Yun et al.~\cite{yun2023imputation} & Singe-Modality & Generative & N/A & General & 2023 \\  
    & & USTD~\cite{hu2023unifying} & Single-Modality & Generative & N/A & General & 2023 \\
    & & PriSTI~\cite{liu2023pristi} & Single-Modality & Generative & N/A & General & 2023 \\

    \hline
    \multirow{13}{*}{\rotatebox[origin=c]{90}{\textbf{Others}}}
    & \multirow{7}{*}{\rotatebox[origin=c]{0}{Transformer-based}}
     &  AuxMobLCast~\cite{xue2022leveraging} & Single-Modality & Pretrained LLM & Fine-tuning \& Prompt-engineering & Mobility & 2022 \\
    & & LLM-Mob~\cite{wang2023i} & Single-Modality & Pretrained LLM & Prompt-engineering & Mobility & 2023 \\
    & & {TrajCL}~\cite{chang2023contrastive} & Single-Modality & Contrastive & Fine-tuning & Mobility & 2023 \\
    & & GTM~\cite{lin2024gtm} & Single-Modality & Generative & Zero-shot & Mobility & 2024 \\
     & & LAMP~\cite{shi2023language} & Singe-Modality &  Pretrained LLM & Prompt-engineering & Event & 2023 \\
    & & Gunjal \& Durrett~\cite{gunjal2023drafting} & Singe-Modality & Pretrained LLM & Prompt-engineering  & Event & 2023 \\
    & & NYUTron~\cite{jiang2023health} & Singe-Modality & Generative & Fine-tuning & Event(Healthcare) & 2023 \\
    & & GatorTron~\cite{yang2022large} & Singe-Modality & Generative & Fine-tuning & Event(Healthcare ) & 2022\\ 

    \cline{2-8}
    & \multirow{3}{*}{\rotatebox[origin=c]{0}{Non-Transformer-based}}   
    &  MMTEC~\cite{lin2023pretraining} & Single-Modality & Contrastive & Fine-tuning & Mobility & 2022 \\
    & & START~\cite{jiang2022self} & Single-Modality & Hybrid & Fine-tuning & Mobility & 2022 \\
    & & Trembr~\cite{fu2020trembr} & Single-Modality & Generative & Fine-tuning & Mobility & 2020 \\ 

    \cline{2-8}
    & \multirow{2}{*}{\rotatebox[origin=c]{0}{Diffusion-based}}  
    &  TrajGDM~\cite{chu2024simulating} & Singe-Modality & Generative & N/A & Mobility & 2024 \\
    & & DiffTraj~\cite{zhu2024difftraj} & Singe-Modality & Generative & N/A & Mobility & 2024 \\
   
    \bottomrule\bottomrule
    \end{tabular}
    }
    \label{tab:summary}
\end{table*}


\end{document}

%% file: img/taxonomy.tex
\tikzstyle{my-box}=[
    rectangle,
    draw=hidden-draw,
    rounded corners,
    text opacity=1,
    minimum height=1.5em,
    minimum width=5em,
    inner sep=2pt,
    align=center,
    fill opacity=.5,
    line width=0.8pt,
]
\tikzstyle{leaf}=[my-box, minimum height=1.5em,
    fill=hidden-pink!80, text=black, align=left, font=\normalsize,
    inner xsep=2pt,
    inner ysep=4pt,
    line width=0.8pt,
]
\begin{figure*}[!t]
    \centering
    \resizebox{0.87\textwidth}{!}
    {
        \begin{forest}
            forked edges,
            for tree={
                fill=level0!80,
                grow=east,
                reversed=true,
                anchor=base west,
                parent anchor=east,
                child anchor=west,
                base=left,
                font=\large,
                rectangle,
                draw=hidden-draw,
                rounded corners,
                align=left,
                minimum width=4em,
                edge+={darkgray, line width=1pt},
                s sep=3pt,
                inner xsep=2pt,
                inner ysep=3pt,
                line width=0.8pt,
                ver/.style={rotate=90, child anchor=north, parent anchor=south, anchor=center},
            },
            where level=1{text width=9em,font=\normalsize,fill=level1!80,}{},
            where level=2{text width=10em,font=\normalsize,fill=level2!80,}{},
            where level=3{text width=8em,font=\normalsize,fill=level3!60,}{},
            where level=4{text width=10em,font=\normalsize,fill=level4!20,}{},
            where level=5{text width=5em,font=\normalsize,}{},
            [Foundation Models  \\ for Time Series 
                [Standard Time Series
                    [Transformer-based
                        [Pre-trained \\ LLM{,} AM{,} VLM
                            [\textbf{General}: Time-LLM~\cite{jin2023timellm}{,} OFA~\cite{zhou2023one}{,} LLM4TS~\cite{chang2023llm4ts}{,}  PromptCast~\cite{xue2022promptcast}{,} \\   TEMPO~\cite{cao2023tempo}{,} LLMTime~\cite{gruver2023large}{,} Voice2Series~\cite{yang2021voice2series}{,} AutoTimes~\cite{liu2024autotimes}{,} UniTime~\cite{liu2024unitime}, leaf, text width=33em]
                            [\textbf{Finance}: Yu \textit{et al.}~\cite{yu2023temporal}{,} Chen \textit{et al.}~\cite{chen2023chatgpt}{,} Xie \textit{et al.}~\cite{xie2023wall}{,} Wimmer \textit{et al.}~\cite{wimmer2023leveraging}, leaf, text width=33em]
                            [\textbf{Healthcare}: Liu \textit{et al.}~\cite{liu2023large}, leaf, text width=22em]
                        ]
                        [Self-supervised
                            [Generative
                                [\textbf{General:} PatchTST~\cite{nie2022time}{,} Moirai~\cite{woo2024unified}{,} Lag-Llama~\cite{rasul2023lag}{,} \\ TimeSiam~\cite{dong2024timesiam}{,} Timer~\cite{liu2024timer}{,} Das \textit{et al.}~\cite{das2023decoder}{,} UniTS~\cite{gao2024units}{,} \\ TimeGPT-1~\cite{garza2023timegpt}{,} Chronos~\cite{ansari2024chronos}{,} MTSMAE~\cite{tang2022mtsmae},  leaf, text width=24em ]
                            ]
                            [Contrastive
                                [\textbf{General:} TEST~\cite{sun2023test}{,} TimeCLR~\cite{yeh2023toward}, leaf, text width=24em ]
                                [\textbf{Healthcare:} METS~\cite{li2023frozen}, leaf, text width=24em]
                            ]
                            [Hybrid
                                [\textbf{General:} SimMTM~\cite{dong2023simmtm}, leaf, text width=24em ]
                            ]
                        ]
                        [Fully-supervised
                            [\textbf{General}: TimeXer~\cite{wang2024timexer}{,} UniTS~\cite{gao2024units}, leaf, text width=22em]
                        ]
                    ]
                    [non-Transformer-based \\ (MLP RNN CNN)
                        [Self-supervised
                            [Generative
                                [\textbf{General:}  TSMixer~\cite{ekambaram2023tsmixer}, leaf, text width=22em ]
                            ]
                            [Contrastive
                                [\textbf{General:} TF-C~\cite{zhang2022self} {,} TS2Vec~\cite{yue2022ts2vec} {,}  CLUDA~\cite{ozyurt2022contrastive} , leaf, text width=22em ]
                            ]
                        ]
                        [Fully-supervised
                            [\textbf{General}: TTMs~\cite{ekambaram2024ttms}{,} TimesNet~\cite{wu2022timesnet}{,} RWKV-TS~\cite{hou2024rwkv}, leaf, text width=25em]
                        ]
                    ]
                    [Diffusion-based
                        [\textbf{General:} TimeGrad~\cite{rasul2021autoregressive} {,} D$^3$VAE~\cite{li2022generative} {,}  TransFusion~\cite{sikder2023transfusion} {,}  ScoreGrad~\cite{yan2021scoregrad} {,} \\  Biloš \textit{et al.}~\cite{bilovs2022modeling} {,}  Crabbé \textit{et al.}~\cite{crabbe2024time} {,}  TimeDiff~\cite{shen2023non} {,}  Wang \textit{et al.} ~\cite{wang2023observed} {,}   DiffTime~\cite{coletta2024constrained}, leaf, text width=35em ]
                        [\textbf{Finance:} {FTS-Diffusion}~\cite{huang2024generative}, leaf, text width=31.7em ]
                        [\textbf{Power:} {DiffLoad}~\cite{wang2023diffload}, leaf, text width=31.7em ]
                    ]
                ]
                [Spatial Time Series
                    [Transformer-based
                        [Pre-trained LLM
                            [\textbf{Transportation:} ST-LLM~\cite{liu2024spatial}{,} TPLLM~\cite{ren2024tpllm}, leaf, text width=22em ]
                            [\textbf{General:} GATGPT~\cite{chen2023gatgpt}, leaf, text width=22em ]
                        ]
                        [Self-supervised
                            [Generative
                                [\textbf{Transportation:} STEP~\cite{shao2022pre}, leaf, text width=22em]
                                [\textbf{Climate:} W-MAE~\cite{man2023w}{,} MetePFL~\cite{chen2023prompt}{,}  FengWu~\cite{chen2023fengwu}, leaf, text width=22em ]
                                [\textbf{General:} UniST~\cite{yuan2024unist}, leaf, text width=22em]
                            ]
                        ]
                        [Fully-supervised
                            [\textbf{Transportation:} CPPBTR~\cite{duan2019pre} {,} TFM~\cite{wang2023building}, leaf, text width=22em ]
                            [\textbf{Climate:} FourCastNet~\cite{pathak2022fourcastnet} {,} FedWing~\cite{chen2023spatial} {,} Pangu-Weather~\cite{bi2023accurate}{,} ClimaX~\cite{nguyen2023climax}, leaf, text width=33em ]
                        ]
                    ]
                    [non-Transformer-based \\ (MLP RNN CNN)
                        [Self-supervised
                            [\textbf{General:} SPGCL~\cite{li2022mining} {,} STGCL~\cite{liu2022contrastive}, leaf, text width=22em ]
                        ]
                    ]
                    [Diffusion-based
                        [\textbf{General:} {DiffSTG}~\cite{wen2023diffstg} {,} {DSTPP}~\cite{yuan2023spatio} {,} {DYffusion}~\cite{cachay2023dyffusion} {,} {Yun \textit{et al.}}~\cite{yun2023imputation} {,} {USTD}~\cite{hu2023unifying} {,} {PriSTI}~\cite{liu2023pristi}, leaf, text width=42em ]
                    ]
                ]
                [Others
                    [Transformer-based
                        [Pre-trained LLM
                            [\textbf{Mobility:} {AuxMobLCast}~\cite{xue2022leveraging} {,} {LLM-Mob}~\cite{wang2023i}, leaf, text width=22em ]
                            [\textbf{Event:} {LAMP}~\cite{shi2023language} {,} {Gunjal \& Durrett \textit{et al.}}~\cite{gunjal2023drafting}, leaf, text width=22em ]
                        ]
                        [Self-supervised
                            [Generative
                                [\textbf{Mobility:} {GTM}~\cite{lin2024gtm}, leaf, text width=22em]
                                [\textbf{Event:} {NYUTron}~\cite{jiang2023health} {,} {GatorTron}~\cite{yang2022large}, leaf, text width=22em ]
                            ]
                            [ Contrastive  
                                 [\textbf{Mobility:} {TrajCL}~\cite{chang2023contrastive}, leaf, text width=22em]
                            ]
                        ]
                    ]
                    [non-Transformer-based \\ (MLP RNN CNN)
                        [Self-supervised
                            [Generative
                               [\textbf{Mobility:} {Trembr}~\cite{fu2020trembr}, leaf, text width=22em ]
                            ]
                            [Contrastive
                               [\textbf{Mobility:} {MMTEC}~\cite{lin2023pretraining}, leaf, text width=22em ] 
                            ]
                            [Hybrid
                               [\textbf{Mobility:} {START}~\cite{jiang2022self}, leaf, text width=22em ] 
                            ]
                        ]
                    ]
                    [Diffusion-based
                        [\textbf{Mobility:} {TrajGDM}~\cite{chu2024simulating} {,} {DiffTraj}~\cite{zhu2024difftraj}, leaf, text width=22em ]
                    ]
                ]
            ]
        \end{forest}
    }    
\caption{A comprehensive taxonomy of TSFMs, categorized according to data and methodologies.}
\label{fig:taxonomy}
\end{figure*}

%

%% file: arxiv.bbl

\begin{thebibliography}{123}


\ifx \showCODEN    \undefined \def \showCODEN     #1{\unskip}     \fi
\ifx \showDOI      \undefined \def \showDOI       #1{#1}\fi
\ifx \showISBNx    \undefined \def \showISBNx     #1{\unskip}     \fi
\ifx \showISBNxiii \undefined \def \showISBNxiii  #1{\unskip}     \fi
\ifx \showISSN     \undefined \def \showISSN      #1{\unskip}     \fi
\ifx \showLCCN     \undefined \def \showLCCN      #1{\unskip}     \fi
\ifx \shownote     \undefined \def \shownote      #1{#1}          \fi
\ifx \showarticletitle \undefined \def \showarticletitle #1{#1}   \fi
\ifx \showURL      \undefined \def \showURL       {\relax}        \fi
\providecommand\bibfield[2]{#2}
\providecommand\bibinfo[2]{#2}
\providecommand\natexlab[1]{#1}
\providecommand\showeprint[2][]{arXiv:#2}

\bibitem[Ansari et~al\mbox{.}(2024)]%
        {ansari2024chronos}
\bibfield{author}{\bibinfo{person}{Abdul~Fatir Ansari}, \bibinfo{person}{Lorenzo Stella}, \bibinfo{person}{Caner Turkmen}, \bibinfo{person}{Xiyuan Zhang}, \bibinfo{person}{Pedro Mercado}, \bibinfo{person}{Huibin Shen}, \bibinfo{person}{Oleksandr Shchur}, \bibinfo{person}{Syama~Syndar Rangapuram}, \bibinfo{person}{Sebastian Pineda~Arango}, \bibinfo{person}{Shubham Kapoor}, \bibinfo{person}{Jasper Zschiegner}, \bibinfo{person}{Danielle~C. Maddix}, \bibinfo{person}{Michael~W. Mahoney}, \bibinfo{person}{Kari Torkkola}, \bibinfo{person}{Andrew Gordon~Wilson}, \bibinfo{person}{Michael Bohlke-Schneider}, {and} \bibinfo{person}{Yuyang Wang}.} \bibinfo{year}{2024}\natexlab{}.
\newblock \showarticletitle{Chronos: Learning the Language of Time Series}.
\newblock \bibinfo{journal}{\emph{arXiv preprint arXiv:2403.07815}} (\bibinfo{year}{2024}).
\newblock


\bibitem[Awais et~al\mbox{.}(2023)]%
        {awais2023foundational}
\bibfield{author}{\bibinfo{person}{Muhammad Awais}, \bibinfo{person}{Muzammal Naseer}, \bibinfo{person}{Salman Khan}, \bibinfo{person}{Rao~Muhammad Anwer}, \bibinfo{person}{Hisham Cholakkal}, \bibinfo{person}{Mubarak Shah}, \bibinfo{person}{Ming-Hsuan Yang}, {and} \bibinfo{person}{Fahad~Shahbaz Khan}.} \bibinfo{year}{2023}\natexlab{}.
\newblock \showarticletitle{Foundational Models Defining a New Era in Vision: A Survey and Outlook}.
\newblock \bibinfo{journal}{\emph{arXiv preprint arXiv:2307.13721}} (\bibinfo{year}{2023}).
\newblock


\bibitem[Bi et~al\mbox{.}(2023a)]%
        {bi2023accurate}
\bibfield{author}{\bibinfo{person}{Kaifeng Bi}, \bibinfo{person}{Lingxi Xie}, \bibinfo{person}{Hengheng Zhang}, \bibinfo{person}{Xin Chen}, \bibinfo{person}{Xiaotao Gu}, {and} \bibinfo{person}{Qi Tian}.} \bibinfo{year}{2023}\natexlab{a}.
\newblock \showarticletitle{Accurate medium-range global weather forecasting with 3D neural networks}.
\newblock \bibinfo{journal}{\emph{Nature}} (\bibinfo{year}{2023}), \bibinfo{pages}{1--6}.
\newblock


\bibitem[Bi et~al\mbox{.}(2023b)]%
        {bi2022pangu}
\bibfield{author}{\bibinfo{person}{Kaifeng Bi}, \bibinfo{person}{Lingxi Xie}, \bibinfo{person}{Hengheng Zhang}, \bibinfo{person}{Xin Chen}, \bibinfo{person}{Xiaotao Gu}, {and} \bibinfo{person}{Qi Tian}.} \bibinfo{year}{2023}\natexlab{b}.
\newblock \showarticletitle{Accurate medium-range global weather forecasting with 3D neural networks}.
\newblock \bibinfo{journal}{\emph{Nature}} \bibinfo{volume}{619}, \bibinfo{number}{7970} (\bibinfo{year}{2023}), \bibinfo{pages}{533--538}.
\newblock


\bibitem[Bilo{\v{s}} et~al\mbox{.}(2022)]%
        {bilovs2022modeling}
\bibfield{author}{\bibinfo{person}{Marin Bilo{\v{s}}}, \bibinfo{person}{Kashif Rasul}, \bibinfo{person}{Anderson Schneider}, \bibinfo{person}{Yuriy Nevmyvaka}, {and} \bibinfo{person}{Stephan G{\"u}nnemann}.} \bibinfo{year}{2022}\natexlab{}.
\newblock \showarticletitle{Modeling temporal data as continuous functions with process diffusion}.
\newblock \bibinfo{journal}{\emph{arXiv preprint arXiv:2211.02590}} (\bibinfo{year}{2022}).
\newblock


\bibitem[Bommasani et~al\mbox{.}(2021)]%
        {bommasani2021opportunities}
\bibfield{author}{\bibinfo{person}{Rishi Bommasani}, \bibinfo{person}{Drew~A Hudson}, \bibinfo{person}{Ehsan Adeli}, \bibinfo{person}{Russ Altman}, \bibinfo{person}{Simran Arora}, \bibinfo{person}{Sydney von Arx}, \bibinfo{person}{Michael~S Bernstein}, \bibinfo{person}{Jeannette Bohg}, \bibinfo{person}{Antoine Bosselut}, \bibinfo{person}{Emma Brunskill}, {et~al\mbox{.}}} \bibinfo{year}{2021}\natexlab{}.
\newblock \showarticletitle{On the opportunities and risks of foundation models}.
\newblock \bibinfo{journal}{\emph{arXiv preprint arXiv:2108.07258}} (\bibinfo{year}{2021}).
\newblock


\bibitem[Brooks et~al\mbox{.}(2024)]%
        {videoworldsimulators2024}
\bibfield{author}{\bibinfo{person}{Tim Brooks}, \bibinfo{person}{Bill Peebles}, \bibinfo{person}{Connor Holmes}, \bibinfo{person}{Will DePue}, \bibinfo{person}{Yufei Guo}, \bibinfo{person}{Li Jing}, \bibinfo{person}{David Schnurr}, \bibinfo{person}{Joe Taylor}, \bibinfo{person}{Troy Luhman}, \bibinfo{person}{Eric Luhman}, \bibinfo{person}{Clarence Ng}, \bibinfo{person}{Ricky Wang}, {and} \bibinfo{person}{Aditya Ramesh}.} \bibinfo{year}{2024}\natexlab{}.
\newblock \showarticletitle{Video generation models as world simulators}.
\newblock  (\bibinfo{year}{2024}).
\newblock
\urldef\tempurl%
\url{https://openai.com/research/video-generation-models-as-world-simulators}
\showURL{%
\tempurl}


\bibitem[Brown et~al\mbox{.}(2020)]%
        {brown2020language}
\bibfield{author}{\bibinfo{person}{Tom Brown}, \bibinfo{person}{Benjamin Mann}, \bibinfo{person}{Nick Ryder}, \bibinfo{person}{Melanie Subbiah}, \bibinfo{person}{Jared~D Kaplan}, \bibinfo{person}{Prafulla Dhariwal}, \bibinfo{person}{Arvind Neelakantan}, \bibinfo{person}{Pranav Shyam}, \bibinfo{person}{Girish Sastry}, \bibinfo{person}{Amanda Askell}, {et~al\mbox{.}}} \bibinfo{year}{2020}\natexlab{}.
\newblock \showarticletitle{Language models are few-shot learners}.
\newblock \bibinfo{journal}{\emph{Advances in neural information processing systems}}  \bibinfo{volume}{33} (\bibinfo{year}{2020}), \bibinfo{pages}{1877--1901}.
\newblock


\bibitem[Cachay et~al\mbox{.}(2023)]%
        {cachay2023dyffusion}
\bibfield{author}{\bibinfo{person}{Salva~R{\"u}hling Cachay}, \bibinfo{person}{Bo Zhao}, \bibinfo{person}{Hailey James}, {and} \bibinfo{person}{Rose Yu}.} \bibinfo{year}{2023}\natexlab{}.
\newblock \showarticletitle{DYffusion: A Dynamics-informed Diffusion Model for Spatiotemporal Forecasting}.
\newblock \bibinfo{journal}{\emph{arXiv preprint arXiv:2306.01984}} (\bibinfo{year}{2023}).
\newblock


\bibitem[Cao et~al\mbox{.}(2023)]%
        {cao2023tempo}
\bibfield{author}{\bibinfo{person}{Defu Cao}, \bibinfo{person}{Furong Jia}, \bibinfo{person}{Sercan~O Arik}, \bibinfo{person}{Tomas Pfister}, \bibinfo{person}{Yixiang Zheng}, \bibinfo{person}{Wen Ye}, {and} \bibinfo{person}{Yan Liu}.} \bibinfo{year}{2023}\natexlab{}.
\newblock \showarticletitle{{TEMPO}: Prompt-based Generative Pre-trained Transformer for Time Series Forecasting}.
\newblock \bibinfo{journal}{\emph{arXiv preprint arXiv:2310.04948}} (\bibinfo{year}{2023}).
\newblock


\bibitem[Chang et~al\mbox{.}(2023a)]%
        {chang2023llm4ts}
\bibfield{author}{\bibinfo{person}{Ching Chang}, \bibinfo{person}{Wen-Chih Peng}, {and} \bibinfo{person}{Tien-Fu Chen}.} \bibinfo{year}{2023}\natexlab{a}.
\newblock \showarticletitle{LLM4TS: Two-Stage Fine-Tuning for Time-Series Forecasting with Pre-Trained LLMs}.
\newblock \bibinfo{journal}{\emph{arXiv preprint arXiv:2308.08469}} (\bibinfo{year}{2023}).
\newblock


\bibitem[Chang et~al\mbox{.}(2023b)]%
        {chang2023contrastive}
\bibfield{author}{\bibinfo{person}{Yanchuan Chang}, \bibinfo{person}{Jianzhong Qi}, \bibinfo{person}{Yuxuan Liang}, {and} \bibinfo{person}{Egemen Tanin}.} \bibinfo{year}{2023}\natexlab{b}.
\newblock \showarticletitle{Contrastive Trajectory Similarity Learning with Dual-Feature Attention}. In \bibinfo{booktitle}{\emph{2023 IEEE 39th International Conference on Data Engineering (ICDE)}}. IEEE, \bibinfo{pages}{2933--2945}.
\newblock


\bibitem[Chen et~al\mbox{.}(2023a)]%
        {chen2023fengwu}
\bibfield{author}{\bibinfo{person}{Kang Chen}, \bibinfo{person}{Tao Han}, \bibinfo{person}{Junchao Gong}, \bibinfo{person}{Lei Bai}, \bibinfo{person}{Fenghua Ling}, \bibinfo{person}{Jing-Jia Luo}, \bibinfo{person}{Xi Chen}, \bibinfo{person}{Leiming Ma}, \bibinfo{person}{Tianning Zhang}, \bibinfo{person}{Rui Su}, {et~al\mbox{.}}} \bibinfo{year}{2023}\natexlab{a}.
\newblock \showarticletitle{FengWu: Pushing the Skillful Global Medium-range Weather Forecast beyond 10 Days Lead}.
\newblock \bibinfo{journal}{\emph{arXiv preprint arXiv:2304.02948}} (\bibinfo{year}{2023}).
\newblock


\bibitem[Chen et~al\mbox{.}(2023b)]%
        {chen2023prompt}
\bibfield{author}{\bibinfo{person}{Shengchao Chen}, \bibinfo{person}{Guodong Long}, \bibinfo{person}{Tao Shen}, {and} \bibinfo{person}{Jing Jiang}.} \bibinfo{year}{2023}\natexlab{b}.
\newblock \showarticletitle{Prompt Federated Learning for Weather Forecasting: Toward Foundation Models on Meteorological Data}. In \bibinfo{booktitle}{\emph{International Joint Conference on Artificial Intelligence}}.
\newblock


\bibitem[Chen et~al\mbox{.}(2023c)]%
        {chen2023spatial}
\bibfield{author}{\bibinfo{person}{Shengchao Chen}, \bibinfo{person}{Guodong Long}, \bibinfo{person}{Tao Shen}, \bibinfo{person}{Tianyi Zhou}, {and} \bibinfo{person}{Jing Jiang}.} \bibinfo{year}{2023}\natexlab{c}.
\newblock \showarticletitle{Spatial-temporal Prompt Learning for Federated Weather Forecasting}.
\newblock \bibinfo{journal}{\emph{arXiv preprint arXiv:2305.14244}} (\bibinfo{year}{2023}).
\newblock


\bibitem[Chen et~al\mbox{.}(2023d)]%
        {chen2023spatialtemporal}
\bibfield{author}{\bibinfo{person}{Shengchao Chen}, \bibinfo{person}{Guodong Long}, \bibinfo{person}{Tao Shen}, \bibinfo{person}{Tianyi Zhou}, {and} \bibinfo{person}{Jing Jiang}.} \bibinfo{year}{2023}\natexlab{d}.
\newblock \bibinfo{title}{Spatial-temporal Prompt Learning for Federated Weather Forecasting}.
\newblock
\newblock
\showeprint[arxiv]{2305.14244}~[cs.LG]


\bibitem[Chen et~al\mbox{.}(2020)]%
        {DBLP:conf/icml/ChenK0H20}
\bibfield{author}{\bibinfo{person}{Ting Chen}, \bibinfo{person}{Simon Kornblith}, \bibinfo{person}{Mohammad Norouzi}, {and} \bibinfo{person}{Geoffrey~E. Hinton}.} \bibinfo{year}{2020}\natexlab{}.
\newblock \showarticletitle{A Simple Framework for Contrastive Learning of Visual Representations}. In \bibinfo{booktitle}{\emph{ICML}}, Vol.~\bibinfo{volume}{119}. \bibinfo{pages}{1597--1607}.
\newblock


\bibitem[Chen et~al\mbox{.}(2023e)]%
        {chen2023gatgpt}
\bibfield{author}{\bibinfo{person}{Yakun Chen}, \bibinfo{person}{Xianzhi Wang}, {and} \bibinfo{person}{Guandong Xu}.} \bibinfo{year}{2023}\natexlab{e}.
\newblock \showarticletitle{Gatgpt: A pre-trained large language model with graph attention network for spatiotemporal imputation}.
\newblock \bibinfo{journal}{\emph{arXiv preprint arXiv:2311.14332}} (\bibinfo{year}{2023}).
\newblock


\bibitem[Chen et~al\mbox{.}(2023f)]%
        {chen2023chatgpt}
\bibfield{author}{\bibinfo{person}{Zihan Chen}, \bibinfo{person}{Lei~Nico Zheng}, \bibinfo{person}{Cheng Lu}, \bibinfo{person}{Jialu Yuan}, {and} \bibinfo{person}{Di Zhu}.} \bibinfo{year}{2023}\natexlab{f}.
\newblock \showarticletitle{ChatGPT Informed Graph Neural Network for Stock Movement Prediction}.
\newblock \bibinfo{journal}{\emph{arXiv preprint arXiv:2306.03763}} (\bibinfo{year}{2023}).
\newblock


\bibitem[Cheng et~al\mbox{.}(2024)]%
        {cheng2024nuwats}
\bibfield{author}{\bibinfo{person}{Jinguo Cheng}, \bibinfo{person}{Chunwei Yang}, \bibinfo{person}{Wanlin Cai}, \bibinfo{person}{Yuxuan Liang}, {and} \bibinfo{person}{Yuankai Wu}.} \bibinfo{year}{2024}\natexlab{}.
\newblock \showarticletitle{NuwaTS: Mending Every Incomplete Time Series}.
\newblock \bibinfo{journal}{\emph{arXiv preprint arXiv:2405.15317}} (\bibinfo{year}{2024}).
\newblock


\bibitem[Chu et~al\mbox{.}(2024)]%
        {chu2024simulating}
\bibfield{author}{\bibinfo{person}{Chen Chu}, \bibinfo{person}{Hengcai Zhang}, \bibinfo{person}{Peixiao Wang}, {and} \bibinfo{person}{Feng Lu}.} \bibinfo{year}{2024}\natexlab{}.
\newblock \showarticletitle{Simulating human mobility with a trajectory generation framework based on diffusion model}.
\newblock \bibinfo{journal}{\emph{International Journal of Geographical Information Science}} (\bibinfo{year}{2024}), \bibinfo{pages}{1--32}.
\newblock


\bibitem[Coletta et~al\mbox{.}(2024)]%
        {coletta2024constrained}
\bibfield{author}{\bibinfo{person}{Andrea Coletta}, \bibinfo{person}{Sriram Gopalakrishnan}, \bibinfo{person}{Daniel Borrajo}, {and} \bibinfo{person}{Svitlana Vyetrenko}.} \bibinfo{year}{2024}\natexlab{}.
\newblock \showarticletitle{On the constrained time-series generation problem}.
\newblock \bibinfo{journal}{\emph{Advances in Neural Information Processing Systems}}  \bibinfo{volume}{36} (\bibinfo{year}{2024}).
\newblock


\bibitem[Crabb{\'e} et~al\mbox{.}(2024)]%
        {crabbe2024time}
\bibfield{author}{\bibinfo{person}{Jonathan Crabb{\'e}}, \bibinfo{person}{Nicolas Huynh}, \bibinfo{person}{Jan Stanczuk}, {and} \bibinfo{person}{Mihaela van~der Schaar}.} \bibinfo{year}{2024}\natexlab{}.
\newblock \showarticletitle{Time Series Diffusion in the Frequency Domain}.
\newblock \bibinfo{journal}{\emph{arXiv preprint arXiv:2402.05933}} (\bibinfo{year}{2024}).
\newblock


\bibitem[Das et~al\mbox{.}(2023)]%
        {das2023decoder}
\bibfield{author}{\bibinfo{person}{Abhimanyu Das}, \bibinfo{person}{Weihao Kong}, \bibinfo{person}{Rajat Sen}, {and} \bibinfo{person}{Yichen Zhou}.} \bibinfo{year}{2023}\natexlab{}.
\newblock \showarticletitle{A decoder-only foundation model for time-series forecasting}.
\newblock \bibinfo{journal}{\emph{arXiv preprint arXiv:2310.10688}} (\bibinfo{year}{2023}).
\newblock


\bibitem[Devlin et~al\mbox{.}(2019)]%
        {DBLP:conf/naacl/DevlinCLT19}
\bibfield{author}{\bibinfo{person}{Jacob Devlin}, \bibinfo{person}{Ming{-}Wei Chang}, \bibinfo{person}{Kenton Lee}, {and} \bibinfo{person}{Kristina Toutanova}.} \bibinfo{year}{2019}\natexlab{}.
\newblock \showarticletitle{{BERT:} Pre-training of Deep Bidirectional Transformers for Language Understanding}. In \bibinfo{booktitle}{\emph{NAACL-HLT}}. \bibinfo{pages}{4171--4186}.
\newblock


\bibitem[Dong et~al\mbox{.}(2024)]%
        {dong2024timesiam}
\bibfield{author}{\bibinfo{person}{Jiaxiang Dong}, \bibinfo{person}{Haixu Wu}, \bibinfo{person}{Yuxuan Wang}, \bibinfo{person}{Yunzhong Qiu}, \bibinfo{person}{Li Zhang}, \bibinfo{person}{Jianmin Wang}, {and} \bibinfo{person}{Mingsheng Long}.} \bibinfo{year}{2024}\natexlab{}.
\newblock \showarticletitle{TimeSiam: A Pre-Training Framework for Siamese Time-Series Modeling}.
\newblock \bibinfo{journal}{\emph{arXiv preprint arXiv:2402.02475}} (\bibinfo{year}{2024}).
\newblock


\bibitem[Dong et~al\mbox{.}(2023)]%
        {dong2023simmtm}
\bibfield{author}{\bibinfo{person}{Jiaxiang Dong}, \bibinfo{person}{Haixu Wu}, \bibinfo{person}{Haoran Zhang}, \bibinfo{person}{Li Zhang}, \bibinfo{person}{Jianmin Wang}, {and} \bibinfo{person}{Mingsheng Long}.} \bibinfo{year}{2023}\natexlab{}.
\newblock \showarticletitle{SimMTM: A Simple Pre-Training Framework for Masked Time-Series Modeling}.
\newblock \bibinfo{journal}{\emph{Advances in Neural Information Processing Systems}} (\bibinfo{year}{2023}).
\newblock


\bibitem[Du et~al\mbox{.}(2021)]%
        {adarnn}
\bibfield{author}{\bibinfo{person}{Yuntao Du}, \bibinfo{person}{Jindong Wang}, \bibinfo{person}{Wenjie Feng}, \bibinfo{person}{Sinno Pan}, \bibinfo{person}{Tao Qin}, \bibinfo{person}{Renjun Xu}, {and} \bibinfo{person}{Chongjun Wang}.} \bibinfo{year}{2021}\natexlab{}.
\newblock \showarticletitle{Adarnn: Adaptive learning and forecasting of time series}. In \bibinfo{booktitle}{\emph{Proceedings of the 30th ACM international conference on information \& knowledge management}}. \bibinfo{pages}{402--411}.
\newblock


\bibitem[Duan et~al\mbox{.}(2019)]%
        {duan2019pre}
\bibfield{author}{\bibinfo{person}{Wenying Duan}, \bibinfo{person}{Liu Jiang}, \bibinfo{person}{Ning Wang}, {and} \bibinfo{person}{Hong Rao}.} \bibinfo{year}{2019}\natexlab{}.
\newblock \showarticletitle{Pre-Trained Bidirectional Temporal Representation for Crowd Flows Prediction in Regular Region}.
\newblock \bibinfo{journal}{\emph{IEEE Access}}  \bibinfo{volume}{7} (\bibinfo{year}{2019}), \bibinfo{pages}{143855--143865}.
\newblock


\bibitem[Ekambaram et~al\mbox{.}(2023)]%
        {ekambaram2023tsmixer}
\bibfield{author}{\bibinfo{person}{Vijay Ekambaram}, \bibinfo{person}{Arindam Jati}, \bibinfo{person}{Nam Nguyen}, \bibinfo{person}{Phanwadee Sinthong}, {and} \bibinfo{person}{Jayant Kalagnanam}.} \bibinfo{year}{2023}\natexlab{}.
\newblock \showarticletitle{TSMixer: Lightweight MLP-Mixer Model for Multivariate Time Series Forecasting}.
\newblock \bibinfo{journal}{\emph{arXiv preprint arXiv:2306.09364}} (\bibinfo{year}{2023}).
\newblock


\bibitem[Ekambaram et~al\mbox{.}(2024)]%
        {ekambaram2024ttms}
\bibfield{author}{\bibinfo{person}{Vijay Ekambaram}, \bibinfo{person}{Arindam Jati}, \bibinfo{person}{Nam~H Nguyen}, \bibinfo{person}{Pankaj Dayama}, \bibinfo{person}{Chandra Reddy}, \bibinfo{person}{Wesley~M Gifford}, {and} \bibinfo{person}{Jayant Kalagnanam}.} \bibinfo{year}{2024}\natexlab{}.
\newblock \showarticletitle{TTMs: Fast Multi-level Tiny Time Mixers for Improved Zero-shot and Few-shot Forecasting of Multivariate Time Series}.
\newblock \bibinfo{journal}{\emph{arXiv preprint arXiv:2401.03955}} (\bibinfo{year}{2024}).
\newblock


\bibitem[Feng et~al\mbox{.}(2024)]%
        {feng2024curve}
\bibfield{author}{\bibinfo{person}{Cheng Feng}, \bibinfo{person}{Long Huang}, {and} \bibinfo{person}{Denis Krompass}.} \bibinfo{year}{2024}\natexlab{}.
\newblock \bibinfo{title}{Only the Curve Shape Matters: Training Foundation Models for Zero-Shot Multivariate Time Series Forecasting through Next Curve Shape Prediction}.
\newblock
\newblock
\showeprint[arxiv]{2402.07570}~[cs.LG]


\bibitem[Fu and Lee(2020)]%
        {fu2020trembr}
\bibfield{author}{\bibinfo{person}{Tao{-}Yang Fu} {and} \bibinfo{person}{Wang{-}Chien Lee}.} \bibinfo{year}{2020}\natexlab{}.
\newblock \showarticletitle{Trembr: Exploring Road Networks for Trajectory Representation Learning}.
\newblock \bibinfo{journal}{\emph{ACM Trans. Intell. Syst. Technol.}} \bibinfo{volume}{11}, \bibinfo{number}{1} (\bibinfo{year}{2020}), \bibinfo{pages}{10:1--10:25}.
\newblock


\bibitem[Gao et~al\mbox{.}(2024)]%
        {gao2024units}
\bibfield{author}{\bibinfo{person}{Shanghua Gao}, \bibinfo{person}{Teddy Koker}, \bibinfo{person}{Owen Queen}, \bibinfo{person}{Thomas Hartvigsen}, \bibinfo{person}{Theodoros Tsiligkaridis}, {and} \bibinfo{person}{Marinka Zitnik}.} \bibinfo{year}{2024}\natexlab{}.
\newblock \showarticletitle{UniTS: Building a Unified Time Series Model}.
\newblock \bibinfo{journal}{\emph{arXiv preprint arXiv:2403.00131}} (\bibinfo{year}{2024}).
\newblock


\bibitem[Garza and Mergenthaler-Canseco(2023)]%
        {garza2023timegpt}
\bibfield{author}{\bibinfo{person}{Azul Garza} {and} \bibinfo{person}{Max Mergenthaler-Canseco}.} \bibinfo{year}{2023}\natexlab{}.
\newblock \showarticletitle{TimeGPT-1}.
\newblock \bibinfo{journal}{\emph{arXiv preprint arXiv:2310.03589}} (\bibinfo{year}{2023}).
\newblock


\bibitem[Gruver et~al\mbox{.}(2023)]%
        {gruver2023large}
\bibfield{author}{\bibinfo{person}{Nate Gruver}, \bibinfo{person}{Marc Finzi}, \bibinfo{person}{Shikai Qiu}, {and} \bibinfo{person}{Andrew~Gordon Wilson}.} \bibinfo{year}{2023}\natexlab{}.
\newblock \showarticletitle{Large Language Models Are Zero-Shot Time Series Forecasters}.
\newblock \bibinfo{journal}{\emph{Advances in Neural Information Processing Systems}} (\bibinfo{year}{2023}).
\newblock


\bibitem[Gu and Dao(2023)]%
        {gu2023mamba}
\bibfield{author}{\bibinfo{person}{Albert Gu} {and} \bibinfo{person}{Tri Dao}.} \bibinfo{year}{2023}\natexlab{}.
\newblock \bibinfo{title}{Mamba: Linear-Time Sequence Modeling with Selective State Spaces}.
\newblock
\newblock
\showeprint[arxiv]{2312.00752}~[cs.LG]


\bibitem[Gunjal and Durrett(2023)]%
        {gunjal2023drafting}
\bibfield{author}{\bibinfo{person}{Anisha Gunjal} {and} \bibinfo{person}{Greg Durrett}.} \bibinfo{year}{2023}\natexlab{}.
\newblock \showarticletitle{Drafting Event Schemas using Language Models}.
\newblock \bibinfo{journal}{\emph{arXiv preprint arXiv:2305.14847}} (\bibinfo{year}{2023}).
\newblock


\bibitem[Hewamalage et~al\mbox{.}(2021)]%
        {hewamalage2021recurrent}
\bibfield{author}{\bibinfo{person}{Hansika Hewamalage}, \bibinfo{person}{Christoph Bergmeir}, {and} \bibinfo{person}{Kasun Bandara}.} \bibinfo{year}{2021}\natexlab{}.
\newblock \showarticletitle{Recurrent neural networks for time series forecasting: Current status and future directions}.
\newblock \bibinfo{journal}{\emph{International Journal of Forecasting}} \bibinfo{volume}{37}, \bibinfo{number}{1} (\bibinfo{year}{2021}), \bibinfo{pages}{388--427}.
\newblock


\bibitem[Hou and Yu(2024)]%
        {hou2024rwkv}
\bibfield{author}{\bibinfo{person}{Haowen Hou} {and} \bibinfo{person}{F~Richard Yu}.} \bibinfo{year}{2024}\natexlab{}.
\newblock \showarticletitle{RWKV-TS: Beyond Traditional Recurrent Neural Network for Time Series Tasks}.
\newblock \bibinfo{journal}{\emph{arXiv preprint arXiv:2401.09093}} (\bibinfo{year}{2024}).
\newblock


\bibitem[Hu et~al\mbox{.}(2023)]%
        {hu2023unifying}
\bibfield{author}{\bibinfo{person}{Junfeng Hu}, \bibinfo{person}{Xu Liu}, \bibinfo{person}{Zhencheng Fan}, \bibinfo{person}{Yuxuan Liang}, {and} \bibinfo{person}{Roger Zimmermann}.} \bibinfo{year}{2023}\natexlab{}.
\newblock \bibinfo{title}{Towards Unifying Diffusion Models for Probabilistic Spatio-Temporal Graph Learning}.
\newblock
\newblock
\showeprint[arxiv]{2310.17360}~[cs.LG]


\bibitem[Huang et~al\mbox{.}(2024)]%
        {huang2024generative}
\bibfield{author}{\bibinfo{person}{Hongbin Huang}, \bibinfo{person}{Minghua Chen}, {and} \bibinfo{person}{Xiao Qiao}.} \bibinfo{year}{2024}\natexlab{}.
\newblock \showarticletitle{Generative Learning for Financial Time Series with Irregular and Scale-Invariant Patterns}. In \bibinfo{booktitle}{\emph{The Twelfth International Conference on Learning Representations}}.
\newblock
\urldef\tempurl%
\url{https://openreview.net/forum?id=CdjnzWsQax}
\showURL{%
\tempurl}


\bibitem[Jiang et~al\mbox{.}(2022)]%
        {jiang2022self}
\bibfield{author}{\bibinfo{person}{Jiawei Jiang}, \bibinfo{person}{Dayan Pan}, \bibinfo{person}{Houxing Ren}, \bibinfo{person}{Xiaohan Jiang}, \bibinfo{person}{Chao Li}, {and} \bibinfo{person}{Jingyuan Wang}.} \bibinfo{year}{2022}\natexlab{}.
\newblock \showarticletitle{Self-supervised Trajectory Representation Learning with Temporal Regularities and Travel Semantics}.
\newblock \bibinfo{journal}{\emph{CoRR}}  \bibinfo{volume}{abs/2211.09510} (\bibinfo{year}{2022}).
\newblock


\bibitem[Jiang et~al\mbox{.}(2023)]%
        {jiang2023health}
\bibfield{author}{\bibinfo{person}{Lavender~Yao Jiang}, \bibinfo{person}{Xujin~Chris Liu}, \bibinfo{person}{Nima~Pour Nejatian}, \bibinfo{person}{Mustafa Nasir-Moin}, \bibinfo{person}{Duo Wang}, \bibinfo{person}{Anas Abidin}, \bibinfo{person}{Kevin Eaton}, \bibinfo{person}{Howard~Antony Riina}, \bibinfo{person}{Ilya Laufer}, \bibinfo{person}{Paawan Punjabi}, {et~al\mbox{.}}} \bibinfo{year}{2023}\natexlab{}.
\newblock \showarticletitle{Health system-scale language models are all-purpose prediction engines}.
\newblock \bibinfo{journal}{\emph{Nature}} (\bibinfo{year}{2023}), \bibinfo{pages}{1--6}.
\newblock


\bibitem[Jiang et~al\mbox{.}(2024)]%
        {jiang2024empowering}
\bibfield{author}{\bibinfo{person}{Yushan Jiang}, \bibinfo{person}{Zijie Pan}, \bibinfo{person}{Xikun Zhang}, \bibinfo{person}{Sahil Garg}, \bibinfo{person}{Anderson Schneider}, \bibinfo{person}{Yuriy Nevmyvaka}, {and} \bibinfo{person}{Dongjin Song}.} \bibinfo{year}{2024}\natexlab{}.
\newblock \showarticletitle{Empowering Time Series Analysis with Large Language Models: A Survey}.
\newblock \bibinfo{journal}{\emph{arXiv preprint arXiv:2402.03182}} (\bibinfo{year}{2024}).
\newblock


\bibitem[Jin et~al\mbox{.}(2023a)]%
        {jin2023survey}
\bibfield{author}{\bibinfo{person}{Ming Jin}, \bibinfo{person}{Huan~Yee Koh}, \bibinfo{person}{Qingsong Wen}, \bibinfo{person}{Daniele Zambon}, \bibinfo{person}{Cesare Alippi}, \bibinfo{person}{Geoffrey~I Webb}, \bibinfo{person}{Irwin King}, {and} \bibinfo{person}{Shirui Pan}.} \bibinfo{year}{2023}\natexlab{a}.
\newblock \showarticletitle{A survey on graph neural networks for time series: Forecasting, classification, imputation, and anomaly detection}.
\newblock \bibinfo{journal}{\emph{arXiv preprint arXiv:2307.03759}} (\bibinfo{year}{2023}).
\newblock


\bibitem[Jin et~al\mbox{.}(2023b)]%
        {jin2023timellm}
\bibfield{author}{\bibinfo{person}{Ming Jin}, \bibinfo{person}{Shiyu Wang}, \bibinfo{person}{Lintao Ma}, \bibinfo{person}{Zhixuan Chu}, \bibinfo{person}{James~Y Zhang}, \bibinfo{person}{Xiaoming Shi}, \bibinfo{person}{Pin-Yu Chen}, \bibinfo{person}{Yuxuan Liang}, \bibinfo{person}{Yuan-Fang Li}, \bibinfo{person}{Shirui Pan}, {et~al\mbox{.}}} \bibinfo{year}{2023}\natexlab{b}.
\newblock \showarticletitle{{Time-LLM}: Time series forecasting by reprogramming large language models}.
\newblock \bibinfo{journal}{\emph{arXiv preprint arXiv:2310.01728}} (\bibinfo{year}{2023}).
\newblock


\bibitem[Jin et~al\mbox{.}(2023c)]%
        {jin2023large}
\bibfield{author}{\bibinfo{person}{Ming Jin}, \bibinfo{person}{Qingsong Wen}, \bibinfo{person}{Yuxuan Liang}, \bibinfo{person}{Chaoli Zhang}, \bibinfo{person}{Siqiao Xue}, \bibinfo{person}{Xue Wang}, \bibinfo{person}{James Zhang}, \bibinfo{person}{Yi Wang}, \bibinfo{person}{Haifeng Chen}, \bibinfo{person}{Xiaoli Li}, {et~al\mbox{.}}} \bibinfo{year}{2023}\natexlab{c}.
\newblock \showarticletitle{Large models for time series and spatio-temporal data: A survey and outlook}.
\newblock \bibinfo{journal}{\emph{arXiv preprint arXiv:2310.10196}} (\bibinfo{year}{2023}).
\newblock


\bibitem[Jin et~al\mbox{.}(2024)]%
        {jin2024position}
\bibfield{author}{\bibinfo{person}{Ming Jin}, \bibinfo{person}{Yifan Zhang}, \bibinfo{person}{Wei Chen}, \bibinfo{person}{Kexin Zhang}, \bibinfo{person}{Yuxuan Liang}, \bibinfo{person}{Bin Yang}, \bibinfo{person}{Jindong Wang}, \bibinfo{person}{Shirui Pan}, {and} \bibinfo{person}{Qingsong Wen}.} \bibinfo{year}{2024}\natexlab{}.
\newblock \showarticletitle{Position: What Can Large Language Models Tell Us about Time Series Analysis}. In \bibinfo{booktitle}{\emph{International Conference on Machine Learning (ICML'24)}}.
\newblock


\bibitem[Kim et~al\mbox{.}(2021)]%
        {kim2021reversible}
\bibfield{author}{\bibinfo{person}{Taesung Kim}, \bibinfo{person}{Jinhee Kim}, \bibinfo{person}{Yunwon Tae}, \bibinfo{person}{Cheonbok Park}, \bibinfo{person}{Jang-Ho Choi}, {and} \bibinfo{person}{Jaegul Choo}.} \bibinfo{year}{2021}\natexlab{}.
\newblock \showarticletitle{Reversible instance normalization for accurate time-series forecasting against distribution shift}. In \bibinfo{booktitle}{\emph{International Conference on Learning Representations}}.
\newblock


\bibitem[Kirillov et~al\mbox{.}(2023)]%
        {kirillov2023segment}
\bibfield{author}{\bibinfo{person}{Alexander Kirillov}, \bibinfo{person}{Eric Mintun}, \bibinfo{person}{Nikhila Ravi}, \bibinfo{person}{Hanzi Mao}, \bibinfo{person}{Chloe Rolland}, \bibinfo{person}{Laura Gustafson}, \bibinfo{person}{Tete Xiao}, \bibinfo{person}{Spencer Whitehead}, \bibinfo{person}{Alexander~C Berg}, \bibinfo{person}{Wan-Yen Lo}, {et~al\mbox{.}}} \bibinfo{year}{2023}\natexlab{}.
\newblock \showarticletitle{Segment anything}. In \bibinfo{booktitle}{\emph{Proceedings of the IEEE/CVF International Conference on Computer Vision}}. \bibinfo{pages}{4015--4026}.
\newblock


\bibitem[Li et~al\mbox{.}(2023)]%
        {li2023frozen}
\bibfield{author}{\bibinfo{person}{Jun Li}, \bibinfo{person}{Che Liu}, \bibinfo{person}{Sibo Cheng}, \bibinfo{person}{Rossella Arcucci}, {and} \bibinfo{person}{Shenda Hong}.} \bibinfo{year}{2023}\natexlab{}.
\newblock \showarticletitle{Frozen Language Model Helps ECG Zero-Shot Learning}. In \bibinfo{booktitle}{\emph{Medical Imaging with Deep Learning}}.
\newblock


\bibitem[Li et~al\mbox{.}(2022b)]%
        {li2022mining}
\bibfield{author}{\bibinfo{person}{Rongfan Li}, \bibinfo{person}{Ting Zhong}, \bibinfo{person}{Xinke Jiang}, \bibinfo{person}{Goce Trajcevski}, \bibinfo{person}{Jin Wu}, {and} \bibinfo{person}{Fan Zhou}.} \bibinfo{year}{2022}\natexlab{b}.
\newblock \showarticletitle{Mining spatio-temporal relations via self-paced graph contrastive learning}. In \bibinfo{booktitle}{\emph{Proceedings of the 28th ACM SIGKDD Conference on Knowledge Discovery and Data Mining}}. \bibinfo{pages}{936--944}.
\newblock


\bibitem[Li et~al\mbox{.}(2019)]%
        {li2019enhancing}
\bibfield{author}{\bibinfo{person}{Shiyang Li}, \bibinfo{person}{Xiaoyong Jin}, \bibinfo{person}{Yao Xuan}, \bibinfo{person}{Xiyou Zhou}, \bibinfo{person}{Wenhu Chen}, \bibinfo{person}{Yu-Xiang Wang}, {and} \bibinfo{person}{Xifeng Yan}.} \bibinfo{year}{2019}\natexlab{}.
\newblock \showarticletitle{Enhancing the locality and breaking the memory bottleneck of transformer on time series forecasting}.
\newblock \bibinfo{journal}{\emph{Advances in neural information processing systems}}  \bibinfo{volume}{32} (\bibinfo{year}{2019}).
\newblock


\bibitem[Li et~al\mbox{.}(2022a)]%
        {li2022generative}
\bibfield{author}{\bibinfo{person}{Yan Li}, \bibinfo{person}{Xinjiang Lu}, \bibinfo{person}{Yaqing Wang}, {and} \bibinfo{person}{Dejing Dou}.} \bibinfo{year}{2022}\natexlab{a}.
\newblock \showarticletitle{Generative time series forecasting with diffusion, denoise, and disentanglement}.
\newblock \bibinfo{journal}{\emph{Advances in Neural Information Processing Systems}}  \bibinfo{volume}{35} (\bibinfo{year}{2022}), \bibinfo{pages}{23009--23022}.
\newblock


\bibitem[Lin et~al\mbox{.}(2024)]%
        {lin2024gtm}
\bibfield{author}{\bibinfo{person}{Yan Lin}, \bibinfo{person}{Jilin Hu}, \bibinfo{person}{Shengnan Guo}, \bibinfo{person}{Bin Yang}, \bibinfo{person}{Christian~S. Jensen}, \bibinfo{person}{Youfang Lin}, {and} \bibinfo{person}{Huaiyu Wan}.} \bibinfo{year}{2024}\natexlab{}.
\newblock \bibinfo{title}{GTM: General Trajectory Modeling with Auto-regressive Generation of Feature Domains}.
\newblock
\newblock
\showeprint[arxiv]{2402.07232}~[cs.LG]


\bibitem[Lin et~al\mbox{.}(2023)]%
        {lin2023pretraining}
\bibfield{author}{\bibinfo{person}{Yan Lin}, \bibinfo{person}{Huaiyu Wan}, \bibinfo{person}{Shengnan Guo}, \bibinfo{person}{Jilin Hu}, \bibinfo{person}{Christian~S. Jensen}, {and} \bibinfo{person}{Youfang Lin}.} \bibinfo{year}{2023}\natexlab{}.
\newblock \bibinfo{title}{Pre-training General Trajectory Embeddings with Maximum Multi-view Entropy Coding}.
\newblock
\newblock
\showeprint[arxiv]{2207.14539}~[cs.CV]


\bibitem[Liu et~al\mbox{.}(2024c)]%
        {liu2024spatial}
\bibfield{author}{\bibinfo{person}{Chenxi Liu}, \bibinfo{person}{Sun Yang}, \bibinfo{person}{Qianxiong Xu}, \bibinfo{person}{Zhishuai Li}, \bibinfo{person}{Cheng Long}, \bibinfo{person}{Ziyue Li}, {and} \bibinfo{person}{Rui Zhao}.} \bibinfo{year}{2024}\natexlab{c}.
\newblock \showarticletitle{Spatial-temporal large language model for traffic prediction}.
\newblock \bibinfo{journal}{\emph{arXiv preprint arXiv:2401.10134}} (\bibinfo{year}{2024}).
\newblock


\bibitem[Liu et~al\mbox{.}(2023a)]%
        {liu2023pristi}
\bibfield{author}{\bibinfo{person}{Mingzhe Liu}, \bibinfo{person}{Han Huang}, \bibinfo{person}{Hao Feng}, \bibinfo{person}{Leilei Sun}, \bibinfo{person}{Bowen Du}, {and} \bibinfo{person}{Yanjie Fu}.} \bibinfo{year}{2023}\natexlab{a}.
\newblock \showarticletitle{PriSTI: A Conditional Diffusion Framework for Spatiotemporal Imputation}.
\newblock \bibinfo{journal}{\emph{arXiv preprint arXiv:2302.09746}} (\bibinfo{year}{2023}).
\newblock


\bibitem[Liu et~al\mbox{.}(2024a)]%
        {liu2024unitime}
\bibfield{author}{\bibinfo{person}{Xu Liu}, \bibinfo{person}{Junfeng Hu}, \bibinfo{person}{Yuan Li}, \bibinfo{person}{Shizhe Diao}, \bibinfo{person}{Yuxuan Liang}, \bibinfo{person}{Bryan Hooi}, {and} \bibinfo{person}{Roger Zimmermann}.} \bibinfo{year}{2024}\natexlab{a}.
\newblock \bibinfo{title}{UniTime: A Language-Empowered Unified Model for Cross-Domain Time Series Forecasting}.
\newblock
\newblock
\showeprint[arxiv]{2310.09751}~[cs.LG]


\bibitem[Liu et~al\mbox{.}(2022)]%
        {liu2022contrastive}
\bibfield{author}{\bibinfo{person}{Xu Liu}, \bibinfo{person}{Yuxuan Liang}, \bibinfo{person}{Chao Huang}, \bibinfo{person}{Yu Zheng}, \bibinfo{person}{Bryan Hooi}, {and} \bibinfo{person}{Roger Zimmermann}.} \bibinfo{year}{2022}\natexlab{}.
\newblock \showarticletitle{When do contrastive learning signals help spatio-temporal graph forecasting?}. In \bibinfo{booktitle}{\emph{Proceedings of the 30th International Conference on Advances in Geographic Information Systems}}. \bibinfo{pages}{1--12}.
\newblock


\bibitem[Liu et~al\mbox{.}(2023b)]%
        {liu2023large}
\bibfield{author}{\bibinfo{person}{Xin Liu}, \bibinfo{person}{Daniel McDuff}, \bibinfo{person}{Geza Kovacs}, \bibinfo{person}{Isaac Galatzer-Levy}, \bibinfo{person}{Jacob Sunshine}, \bibinfo{person}{Jiening Zhan}, \bibinfo{person}{Ming-Zher Poh}, \bibinfo{person}{Shun Liao}, \bibinfo{person}{Paolo Di~Achille}, {and} \bibinfo{person}{Shwetak Patel}.} \bibinfo{year}{2023}\natexlab{b}.
\newblock \showarticletitle{Large Language Models are Few-Shot Health Learners}.
\newblock \bibinfo{journal}{\emph{arXiv preprint arXiv:2305.15525}} (\bibinfo{year}{2023}).
\newblock


\bibitem[Liu et~al\mbox{.}(2024b)]%
        {liu2024autotimes}
\bibfield{author}{\bibinfo{person}{Yong Liu}, \bibinfo{person}{Guo Qin}, \bibinfo{person}{Xiangdong Huang}, \bibinfo{person}{Jianmin Wang}, {and} \bibinfo{person}{Mingsheng Long}.} \bibinfo{year}{2024}\natexlab{b}.
\newblock \showarticletitle{AutoTimes: Autoregressive Time Series Forecasters via Large Language Models}.
\newblock \bibinfo{journal}{\emph{arXiv preprint arXiv:2402.02370}} (\bibinfo{year}{2024}).
\newblock


\bibitem[Liu et~al\mbox{.}(2024d)]%
        {liu2024timer}
\bibfield{author}{\bibinfo{person}{Yong Liu}, \bibinfo{person}{Haoran Zhang}, \bibinfo{person}{Chenyu Li}, \bibinfo{person}{Xiangdong Huang}, \bibinfo{person}{Jianmin Wang}, {and} \bibinfo{person}{Mingsheng Long}.} \bibinfo{year}{2024}\natexlab{d}.
\newblock \showarticletitle{Timer: Transformers for Time Series Analysis at Scale}.
\newblock \bibinfo{journal}{\emph{arXiv preprint arXiv:2402.02368}} (\bibinfo{year}{2024}).
\newblock


\bibitem[Man et~al\mbox{.}(2023)]%
        {man2023w}
\bibfield{author}{\bibinfo{person}{Xin Man}, \bibinfo{person}{Chenghong Zhang}, \bibinfo{person}{Changyu Li}, {and} \bibinfo{person}{Jie Shao}.} \bibinfo{year}{2023}\natexlab{}.
\newblock \showarticletitle{W-MAE: Pre-trained weather model with masked autoencoder for multi-variable weather forecasting}.
\newblock \bibinfo{journal}{\emph{arXiv preprint arXiv:2304.08754}} (\bibinfo{year}{2023}).
\newblock


\bibitem[Miller et~al\mbox{.}(2024)]%
        {miller2024survey}
\bibfield{author}{\bibinfo{person}{John~A. Miller}, \bibinfo{person}{Mohammed Aldosari}, \bibinfo{person}{Farah Saeed}, \bibinfo{person}{Nasid~Habib Barna}, \bibinfo{person}{Subas Rana}, \bibinfo{person}{I.~Budak Arpinar}, {and} \bibinfo{person}{Ninghao Liu}.} \bibinfo{year}{2024}\natexlab{}.
\newblock \bibinfo{title}{A Survey of Deep Learning and Foundation Models for Time Series Forecasting}.
\newblock
\newblock
\showeprint[arxiv]{2401.13912}~[cs.LG]


\bibitem[Nguyen et~al\mbox{.}(2023)]%
        {nguyen2023climax}
\bibfield{author}{\bibinfo{person}{Tung Nguyen}, \bibinfo{person}{Johannes Brandstetter}, \bibinfo{person}{Ashish Kapoor}, \bibinfo{person}{Jayesh~K Gupta}, {and} \bibinfo{person}{Aditya Grover}.} \bibinfo{year}{2023}\natexlab{}.
\newblock \showarticletitle{ClimaX: A foundation model for weather and climate}.
\newblock \bibinfo{journal}{\emph{International Conference on Machine Learning}} (\bibinfo{year}{2023}).
\newblock


\bibitem[Nie et~al\mbox{.}(2022)]%
        {nie2022time}
\bibfield{author}{\bibinfo{person}{Yuqi Nie}, \bibinfo{person}{Nam~H Nguyen}, \bibinfo{person}{Phanwadee Sinthong}, {and} \bibinfo{person}{Jayant Kalagnanam}.} \bibinfo{year}{2022}\natexlab{}.
\newblock \showarticletitle{A time series is worth 64 words: Long-term forecasting with transformers}.
\newblock \bibinfo{journal}{\emph{arXiv preprint arXiv:2211.14730}} (\bibinfo{year}{2022}).
\newblock


\bibitem[Ozyurt et~al\mbox{.}(2022)]%
        {ozyurt2022contrastive}
\bibfield{author}{\bibinfo{person}{Yilmazcan Ozyurt}, \bibinfo{person}{Stefan Feuerriegel}, {and} \bibinfo{person}{Ce Zhang}.} \bibinfo{year}{2022}\natexlab{}.
\newblock \showarticletitle{Contrastive learning for unsupervised domain adaptation of time series}.
\newblock \bibinfo{journal}{\emph{arXiv preprint arXiv:2206.06243}} (\bibinfo{year}{2022}).
\newblock


\bibitem[Pathak et~al\mbox{.}(2022)]%
        {pathak2022fourcastnet}
\bibfield{author}{\bibinfo{person}{Jaideep Pathak}, \bibinfo{person}{Shashank Subramanian}, \bibinfo{person}{Peter Harrington}, \bibinfo{person}{Sanjeev Raja}, \bibinfo{person}{Ashesh Chattopadhyay}, \bibinfo{person}{Morteza Mardani}, \bibinfo{person}{Thorsten Kurth}, \bibinfo{person}{David Hall}, \bibinfo{person}{Zongyi Li}, \bibinfo{person}{Kamyar Azizzadenesheli}, {et~al\mbox{.}}} \bibinfo{year}{2022}\natexlab{}.
\newblock \showarticletitle{Fourcastnet: A global data-driven high-resolution weather model using adaptive fourier neural operators}.
\newblock \bibinfo{journal}{\emph{arXiv preprint arXiv:2202.11214}} (\bibinfo{year}{2022}).
\newblock


\bibitem[Peebles and Xie(2023)]%
        {peebles2023scalable}
\bibfield{author}{\bibinfo{person}{William Peebles} {and} \bibinfo{person}{Saining Xie}.} \bibinfo{year}{2023}\natexlab{}.
\newblock \showarticletitle{Scalable diffusion models with transformers}. In \bibinfo{booktitle}{\emph{Proceedings of the IEEE/CVF International Conference on Computer Vision}}. \bibinfo{pages}{4195--4205}.
\newblock


\bibitem[Peng et~al\mbox{.}(2023)]%
        {peng2023rwkv}
\bibfield{author}{\bibinfo{person}{Bo Peng}, \bibinfo{person}{Eric Alcaide}, \bibinfo{person}{Quentin Anthony}, \bibinfo{person}{Alon Albalak}, \bibinfo{person}{Samuel Arcadinho}, \bibinfo{person}{Huanqi Cao}, \bibinfo{person}{Xin Cheng}, \bibinfo{person}{Michael Chung}, \bibinfo{person}{Matteo Grella}, \bibinfo{person}{Kranthi~Kiran GV}, {et~al\mbox{.}}} \bibinfo{year}{2023}\natexlab{}.
\newblock \showarticletitle{Rwkv: Reinventing rnns for the transformer era}.
\newblock \bibinfo{journal}{\emph{arXiv preprint arXiv:2305.13048}} (\bibinfo{year}{2023}).
\newblock


\bibitem[Radford et~al\mbox{.}(2021)]%
        {radford2021learning}
\bibfield{author}{\bibinfo{person}{Alec Radford}, \bibinfo{person}{Jong~Wook Kim}, \bibinfo{person}{Chris Hallacy}, \bibinfo{person}{Aditya Ramesh}, \bibinfo{person}{Gabriel Goh}, \bibinfo{person}{Sandhini Agarwal}, \bibinfo{person}{Girish Sastry}, \bibinfo{person}{Amanda Askell}, \bibinfo{person}{Pamela Mishkin}, \bibinfo{person}{Jack Clark}, {et~al\mbox{.}}} \bibinfo{year}{2021}\natexlab{}.
\newblock \showarticletitle{Learning transferable visual models from natural language supervision}. In \bibinfo{booktitle}{\emph{International conference on machine learning}}. PMLR, \bibinfo{pages}{8748--8763}.
\newblock


\bibitem[Radford et~al\mbox{.}(2019)]%
        {radford2019language}
\bibfield{author}{\bibinfo{person}{Alec Radford}, \bibinfo{person}{Jeffrey Wu}, \bibinfo{person}{Rewon Child}, \bibinfo{person}{David Luan}, \bibinfo{person}{Dario Amodei}, \bibinfo{person}{Ilya Sutskever}, {et~al\mbox{.}}} \bibinfo{year}{2019}\natexlab{}.
\newblock \showarticletitle{Language models are unsupervised multitask learners}.
\newblock \bibinfo{journal}{\emph{OpenAI blog}} \bibinfo{volume}{1}, \bibinfo{number}{8} (\bibinfo{year}{2019}), \bibinfo{pages}{9}.
\newblock


\bibitem[Rasul et~al\mbox{.}(2023)]%
        {rasul2023lag}
\bibfield{author}{\bibinfo{person}{Kashif Rasul}, \bibinfo{person}{Arjun Ashok}, \bibinfo{person}{Andrew~Robert Williams}, \bibinfo{person}{Arian Khorasani}, \bibinfo{person}{George Adamopoulos}, \bibinfo{person}{Rishika Bhagwatkar}, \bibinfo{person}{Marin Bilo{\v{s}}}, \bibinfo{person}{Hena Ghonia}, \bibinfo{person}{Nadhir~Vincent Hassen}, \bibinfo{person}{Anderson Schneider}, {et~al\mbox{.}}} \bibinfo{year}{2023}\natexlab{}.
\newblock \showarticletitle{Lag-llama: Towards foundation models for time series forecasting}.
\newblock \bibinfo{journal}{\emph{arXiv preprint arXiv:2310.08278}} (\bibinfo{year}{2023}).
\newblock


\bibitem[Rasul et~al\mbox{.}(2021)]%
        {rasul2021autoregressive}
\bibfield{author}{\bibinfo{person}{Kashif Rasul}, \bibinfo{person}{Calvin Seward}, \bibinfo{person}{Ingmar Schuster}, {and} \bibinfo{person}{Roland Vollgraf}.} \bibinfo{year}{2021}\natexlab{}.
\newblock \showarticletitle{Autoregressive denoising diffusion models for multivariate probabilistic time series forecasting}. In \bibinfo{booktitle}{\emph{International Conference on Machine Learning}}. PMLR, \bibinfo{pages}{8857--8868}.
\newblock


\bibitem[Ren et~al\mbox{.}(2024)]%
        {ren2024tpllm}
\bibfield{author}{\bibinfo{person}{Yilong Ren}, \bibinfo{person}{Yue Chen}, \bibinfo{person}{Shuai Liu}, \bibinfo{person}{Boyue Wang}, \bibinfo{person}{Haiyang Yu}, {and} \bibinfo{person}{Zhiyong Cui}.} \bibinfo{year}{2024}\natexlab{}.
\newblock \showarticletitle{TPLLM: A Traffic Prediction Framework Based on Pretrained Large Language Models}.
\newblock \bibinfo{journal}{\emph{arXiv preprint arXiv:2403.02221}} (\bibinfo{year}{2024}).
\newblock


\bibitem[Rombach et~al\mbox{.}(2022)]%
        {rombach2022high}
\bibfield{author}{\bibinfo{person}{Robin Rombach}, \bibinfo{person}{Andreas Blattmann}, \bibinfo{person}{Dominik Lorenz}, \bibinfo{person}{Patrick Esser}, {and} \bibinfo{person}{Bj{\"o}rn Ommer}.} \bibinfo{year}{2022}\natexlab{}.
\newblock \showarticletitle{High-resolution image synthesis with latent diffusion models}. In \bibinfo{booktitle}{\emph{Proceedings of the IEEE/CVF conference on computer vision and pattern recognition}}. \bibinfo{pages}{10684--10695}.
\newblock


\bibitem[Shao et~al\mbox{.}(2022)]%
        {shao2022pre}
\bibfield{author}{\bibinfo{person}{Zezhi Shao}, \bibinfo{person}{Zhao Zhang}, \bibinfo{person}{Fei Wang}, {and} \bibinfo{person}{Yongjun Xu}.} \bibinfo{year}{2022}\natexlab{}.
\newblock \showarticletitle{Pre-training enhanced spatial-temporal graph neural network for multivariate time series forecasting}. In \bibinfo{booktitle}{\emph{Proceedings of the 28th ACM SIGKDD Conference on Knowledge Discovery and Data Mining}}. \bibinfo{pages}{1567--1577}.
\newblock


\bibitem[Shen and Kwok(2023)]%
        {shen2023non}
\bibfield{author}{\bibinfo{person}{Lifeng Shen} {and} \bibinfo{person}{James Kwok}.} \bibinfo{year}{2023}\natexlab{}.
\newblock \showarticletitle{Non-autoregressive Conditional Diffusion Models for Time Series Prediction}.
\newblock \bibinfo{journal}{\emph{arXiv preprint arXiv:2306.05043}} (\bibinfo{year}{2023}).
\newblock


\bibitem[Shi et~al\mbox{.}(2023)]%
        {shi2023language}
\bibfield{author}{\bibinfo{person}{Xiaoming Shi}, \bibinfo{person}{Siqiao Xue}, \bibinfo{person}{Kangrui Wang}, \bibinfo{person}{Fan Zhou}, \bibinfo{person}{James~Y Zhang}, \bibinfo{person}{Jun Zhou}, \bibinfo{person}{Chenhao Tan}, {and} \bibinfo{person}{Hongyuan Mei}.} \bibinfo{year}{2023}\natexlab{}.
\newblock \showarticletitle{Language Models Can Improve Event Prediction by Few-Shot Abductive Reasoning}. In \bibinfo{booktitle}{\emph{Advances in Neural Information Processing Systems}}.
\newblock


\bibitem[Sikder et~al\mbox{.}(2023)]%
        {sikder2023transfusion}
\bibfield{author}{\bibinfo{person}{Md~Fahim Sikder}, \bibinfo{person}{Resmi Ramachandranpillai}, {and} \bibinfo{person}{Fredrik Heintz}.} \bibinfo{year}{2023}\natexlab{}.
\newblock \showarticletitle{Transfusion: generating long, high fidelity time series using diffusion models with transformers}.
\newblock \bibinfo{journal}{\emph{arXiv preprint arXiv:2307.12667}} (\bibinfo{year}{2023}).
\newblock


\bibitem[Sun et~al\mbox{.}(2023)]%
        {sun2023test}
\bibfield{author}{\bibinfo{person}{Chenxi Sun}, \bibinfo{person}{Yaliang Li}, \bibinfo{person}{Hongyan Li}, {and} \bibinfo{person}{Shenda Hong}.} \bibinfo{year}{2023}\natexlab{}.
\newblock \showarticletitle{TEST: Text Prototype Aligned Embedding to Activate LLM's Ability for Time Series}.
\newblock \bibinfo{journal}{\emph{arXiv preprint arXiv:2308.08241}} (\bibinfo{year}{2023}).
\newblock


\bibitem[Tang and Zhang(2022)]%
        {tang2022mtsmae}
\bibfield{author}{\bibinfo{person}{Peiwang Tang} {and} \bibinfo{person}{Xianchao Zhang}.} \bibinfo{year}{2022}\natexlab{}.
\newblock \showarticletitle{MTSMAE: Masked Autoencoders for Multivariate Time-Series Forecasting}. In \bibinfo{booktitle}{\emph{2022 IEEE 34th International Conference on Tools with Artificial Intelligence (ICTAI)}}. IEEE, \bibinfo{pages}{982--989}.
\newblock


\bibitem[Vaswani et~al\mbox{.}(2017)]%
        {vaswani2017transformer}
\bibfield{author}{\bibinfo{person}{Ashish Vaswani}, \bibinfo{person}{Noam Shazeer}, \bibinfo{person}{Niki Parmar}, \bibinfo{person}{Jakob Uszkoreit}, \bibinfo{person}{Llion Jones}, \bibinfo{person}{Aidan~N Gomez}, \bibinfo{person}{{\L}ukasz Kaiser}, {and} \bibinfo{person}{Illia Polosukhin}.} \bibinfo{year}{2017}\natexlab{}.
\newblock \showarticletitle{Attention is all you need}. In \bibinfo{booktitle}{\emph{Advances in Neural Information Processing Systems 30}}. \bibinfo{pages}{5998--6008}.
\newblock


\bibitem[Wang et~al\mbox{.}(2024a)]%
        {wang2024deep}
\bibfield{author}{\bibinfo{person}{Jun Wang}, \bibinfo{person}{Wenjie Du}, \bibinfo{person}{Wei Cao}, \bibinfo{person}{Keli Zhang}, \bibinfo{person}{Wenjia Wang}, \bibinfo{person}{Yuxuan Liang}, {and} \bibinfo{person}{Qingsong Wen}.} \bibinfo{year}{2024}\natexlab{a}.
\newblock \showarticletitle{Deep Learning for Multivariate Time Series Imputation: A Survey}.
\newblock \bibinfo{journal}{\emph{arXiv preprint arXiv:2402.04059}} (\bibinfo{year}{2024}).
\newblock


\bibitem[Wang et~al\mbox{.}(2023a)]%
        {wang2023i}
\bibfield{author}{\bibinfo{person}{Xinglei Wang}, \bibinfo{person}{Meng Fang}, \bibinfo{person}{Zichao Zeng}, {and} \bibinfo{person}{Tao Cheng}.} \bibinfo{year}{2023}\natexlab{a}.
\newblock \bibinfo{title}{Where Would I Go Next? Large Language Models as Human Mobility Predictors}.
\newblock
\newblock
\showeprint[arxiv]{2308.15197}~[cs.AI]


\bibitem[Wang et~al\mbox{.}(2023c)]%
        {wang2023building}
\bibfield{author}{\bibinfo{person}{Xuhong Wang}, \bibinfo{person}{Ding Wang}, \bibinfo{person}{Liang Chen}, {and} \bibinfo{person}{Yilun Lin}.} \bibinfo{year}{2023}\natexlab{c}.
\newblock \showarticletitle{Building Transportation Foundation Model via Generative Graph Transformer}.
\newblock \bibinfo{journal}{\emph{arXiv preprint arXiv:2305.14826}} (\bibinfo{year}{2023}).
\newblock


\bibitem[Wang et~al\mbox{.}(2023e)]%
        {wang2023observed}
\bibfield{author}{\bibinfo{person}{Xu Wang}, \bibinfo{person}{Hongbo Zhang}, \bibinfo{person}{Pengkun Wang}, \bibinfo{person}{Yudong Zhang}, \bibinfo{person}{Binwu Wang}, \bibinfo{person}{Zhengyang Zhou}, {and} \bibinfo{person}{Yang Wang}.} \bibinfo{year}{2023}\natexlab{e}.
\newblock \showarticletitle{An observed value consistent diffusion model for imputing missing values in multivariate time series}. In \bibinfo{booktitle}{\emph{Proceedings of the 29th ACM SIGKDD Conference on Knowledge Discovery and Data Mining}}. \bibinfo{pages}{2409--2418}.
\newblock


\bibitem[Wang et~al\mbox{.}(2024b)]%
        {wang2024timexer}
\bibfield{author}{\bibinfo{person}{Yuxuan Wang}, \bibinfo{person}{Haixu Wu}, \bibinfo{person}{Jiaxiang Dong}, \bibinfo{person}{Yong Liu}, \bibinfo{person}{Yunzhong Qiu}, \bibinfo{person}{Haoran Zhang}, \bibinfo{person}{Jianmin Wang}, {and} \bibinfo{person}{Mingsheng Long}.} \bibinfo{year}{2024}\natexlab{b}.
\newblock \showarticletitle{TimeXer: Empowering Transformers for Time Series Forecasting with Exogenous Variables}.
\newblock \bibinfo{journal}{\emph{arXiv preprint arXiv:2402.19072}} (\bibinfo{year}{2024}).
\newblock


\bibitem[Wang et~al\mbox{.}(2023b)]%
        {wang2023st}
\bibfield{author}{\bibinfo{person}{Zepu Wang}, \bibinfo{person}{Yuqi Nie}, \bibinfo{person}{Peng Sun}, \bibinfo{person}{Nam~H Nguyen}, \bibinfo{person}{John Mulvey}, {and} \bibinfo{person}{H~Vincent Poor}.} \bibinfo{year}{2023}\natexlab{b}.
\newblock \showarticletitle{St-mlp: A cascaded spatio-temporal linear framework with channel-independence strategy for traffic forecasting}.
\newblock \bibinfo{journal}{\emph{arXiv preprint arXiv:2308.07496}} (\bibinfo{year}{2023}).
\newblock


\bibitem[Wang et~al\mbox{.}(2023d)]%
        {wang2023diffload}
\bibfield{author}{\bibinfo{person}{Zhixian Wang}, \bibinfo{person}{Qingsong Wen}, \bibinfo{person}{Chaoli Zhang}, \bibinfo{person}{Liang Sun}, {and} \bibinfo{person}{Yi Wang}.} \bibinfo{year}{2023}\natexlab{d}.
\newblock \showarticletitle{{DiffLoad}: uncertainty quantification in load forecasting with diffusion model}.
\newblock \bibinfo{journal}{\emph{arXiv preprint arXiv:2306.01001}} (\bibinfo{year}{2023}).
\newblock


\bibitem[Wen et~al\mbox{.}(2023a)]%
        {wen2023diffstg}
\bibfield{author}{\bibinfo{person}{Haomin Wen}, \bibinfo{person}{Youfang Lin}, \bibinfo{person}{Yutong Xia}, \bibinfo{person}{Huaiyu Wan}, \bibinfo{person}{Qingsong Wen}, \bibinfo{person}{Roger Zimmermann}, {and} \bibinfo{person}{Yuxuan Liang}.} \bibinfo{year}{2023}\natexlab{a}.
\newblock \showarticletitle{{DiffSTG}: Probabilistic spatio-temporal graph forecasting with denoising diffusion models}. In \bibinfo{booktitle}{\emph{the 31st ACM International Conference on Advances in Geographic Information Systems}}. \bibinfo{pages}{1--12}.
\newblock


\bibitem[Wen et~al\mbox{.}(2022)]%
        {wen2022robust}
\bibfield{author}{\bibinfo{person}{Qingsong Wen}, \bibinfo{person}{Linxiao Yang}, \bibinfo{person}{Tian Zhou}, {and} \bibinfo{person}{Liang Sun}.} \bibinfo{year}{2022}\natexlab{}.
\newblock \showarticletitle{Robust time series analysis and applications: An industrial perspective}. In \bibinfo{booktitle}{\emph{28th ACM SIGKDD Conference on Knowledge Discovery and Data Mining}}. \bibinfo{pages}{4836--4837}.
\newblock


\bibitem[Wen et~al\mbox{.}(2023b)]%
        {wen2022transformers}
\bibfield{author}{\bibinfo{person}{Qingsong Wen}, \bibinfo{person}{Tian Zhou}, \bibinfo{person}{Chaoli Zhang}, \bibinfo{person}{Weiqi Chen}, \bibinfo{person}{Ziqing Ma}, \bibinfo{person}{Junchi Yan}, {and} \bibinfo{person}{Liang Sun}.} \bibinfo{year}{2023}\natexlab{b}.
\newblock \showarticletitle{Transformers in time series: A survey}. In \bibinfo{booktitle}{\emph{International Joint Conference on Artificial Intelligence(IJCAI)}}. \bibinfo{pages}{6778--6786}.
\newblock


\bibitem[Wimmer and Rekabsaz(2023)]%
        {wimmer2023leveraging}
\bibfield{author}{\bibinfo{person}{Christopher Wimmer} {and} \bibinfo{person}{Navid Rekabsaz}.} \bibinfo{year}{2023}\natexlab{}.
\newblock \showarticletitle{Leveraging vision-language models for granular market change prediction}.
\newblock \bibinfo{journal}{\emph{arXiv preprint arXiv:2301.10166}} (\bibinfo{year}{2023}).
\newblock


\bibitem[Woo et~al\mbox{.}(2024)]%
        {woo2024unified}
\bibfield{author}{\bibinfo{person}{Gerald Woo}, \bibinfo{person}{Chenghao Liu}, \bibinfo{person}{Akshat Kumar}, \bibinfo{person}{Caiming Xiong}, \bibinfo{person}{Silvio Savarese}, {and} \bibinfo{person}{Doyen Sahoo}.} \bibinfo{year}{2024}\natexlab{}.
\newblock \showarticletitle{Unified training of universal time series forecasting transformers}.
\newblock \bibinfo{journal}{\emph{arXiv preprint arXiv:2402.02592}} (\bibinfo{year}{2024}).
\newblock


\bibitem[Wu et~al\mbox{.}(2022)]%
        {wu2022timesnet}
\bibfield{author}{\bibinfo{person}{Haixu Wu}, \bibinfo{person}{Tengge Hu}, \bibinfo{person}{Yong Liu}, \bibinfo{person}{Hang Zhou}, \bibinfo{person}{Jianmin Wang}, {and} \bibinfo{person}{Mingsheng Long}.} \bibinfo{year}{2022}\natexlab{}.
\newblock \showarticletitle{Timesnet: Temporal 2d-variation modeling for general time series analysis}. In \bibinfo{booktitle}{\emph{The eleventh international conference on learning representations}}.
\newblock


\bibitem[Xia et~al\mbox{.}(2024)]%
        {xia2024deciphering}
\bibfield{author}{\bibinfo{person}{Yutong Xia}, \bibinfo{person}{Yuxuan Liang}, \bibinfo{person}{Haomin Wen}, \bibinfo{person}{Xu Liu}, \bibinfo{person}{Kun Wang}, \bibinfo{person}{Zhengyang Zhou}, {and} \bibinfo{person}{Roger Zimmermann}.} \bibinfo{year}{2024}\natexlab{}.
\newblock \showarticletitle{Deciphering spatio-temporal graph forecasting: A causal lens and treatment}.
\newblock \bibinfo{journal}{\emph{Advances in Neural Information Processing Systems}}  \bibinfo{volume}{36} (\bibinfo{year}{2024}).
\newblock


\bibitem[Xie et~al\mbox{.}(2023)]%
        {xie2023wall}
\bibfield{author}{\bibinfo{person}{Qianqian Xie}, \bibinfo{person}{Weiguang Han}, \bibinfo{person}{Yanzhao Lai}, \bibinfo{person}{Min Peng}, {and} \bibinfo{person}{Jimin Huang}.} \bibinfo{year}{2023}\natexlab{}.
\newblock \showarticletitle{The Wall Street Neophyte: A Zero-Shot Analysis of ChatGPT Over MultiModal Stock Movement Prediction Challenges}.
\newblock \bibinfo{journal}{\emph{arXiv preprint arXiv:2304.05351}} (\bibinfo{year}{2023}).
\newblock


\bibitem[Xu et~al\mbox{.}(2021)]%
        {xu2021anomaly}
\bibfield{author}{\bibinfo{person}{Jiehui Xu}, \bibinfo{person}{Haixu Wu}, \bibinfo{person}{Jianmin Wang}, {and} \bibinfo{person}{Mingsheng Long}.} \bibinfo{year}{2021}\natexlab{}.
\newblock \showarticletitle{Anomaly Transformer: Time Series Anomaly Detection with Association Discrepancy}. In \bibinfo{booktitle}{\emph{International Conference on Learning Representations}}.
\newblock


\bibitem[Xue and Salim(2022)]%
        {xue2022promptcast}
\bibfield{author}{\bibinfo{person}{Hao Xue} {and} \bibinfo{person}{Flora~D Salim}.} \bibinfo{year}{2022}\natexlab{}.
\newblock \showarticletitle{PromptCast: A New Prompt-based Learning Paradigm for Time Series Forecasting}.
\newblock \bibinfo{journal}{\emph{arXiv preprint arXiv:2210.08964}} (\bibinfo{year}{2022}).
\newblock


\bibitem[Xue et~al\mbox{.}(2022)]%
        {xue2022leveraging}
\bibfield{author}{\bibinfo{person}{Hao Xue}, \bibinfo{person}{Bhanu~Prakash Voutharoja}, {and} \bibinfo{person}{Flora~D Salim}.} \bibinfo{year}{2022}\natexlab{}.
\newblock \showarticletitle{Leveraging language foundation models for human mobility forecasting}. In \bibinfo{booktitle}{\emph{the 30th International Conference on Advances in Geographic Information Systems}}. \bibinfo{pages}{1--9}.
\newblock


\bibitem[Yan et~al\mbox{.}(2021)]%
        {yan2021scoregrad}
\bibfield{author}{\bibinfo{person}{Tijin Yan}, \bibinfo{person}{Hongwei Zhang}, \bibinfo{person}{Tong Zhou}, \bibinfo{person}{Yufeng Zhan}, {and} \bibinfo{person}{Yuanqing Xia}.} \bibinfo{year}{2021}\natexlab{}.
\newblock \showarticletitle{ScoreGrad: Multivariate probabilistic time series forecasting with continuous energy-based generative models}.
\newblock \bibinfo{journal}{\emph{arXiv preprint arXiv:2106.10121}} (\bibinfo{year}{2021}).
\newblock


\bibitem[Yang et~al\mbox{.}(2021)]%
        {yang2021voice2series}
\bibfield{author}{\bibinfo{person}{Chao-Han~Huck Yang}, \bibinfo{person}{Yun-Yun Tsai}, {and} \bibinfo{person}{Pin-Yu Chen}.} \bibinfo{year}{2021}\natexlab{}.
\newblock \showarticletitle{Voice2series: Reprogramming acoustic models for time series classification}. In \bibinfo{booktitle}{\emph{International conference on machine learning}}. PMLR, \bibinfo{pages}{11808--11819}.
\newblock


\bibitem[Yang et~al\mbox{.}(2022)]%
        {yang2022large}
\bibfield{author}{\bibinfo{person}{Xi Yang}, \bibinfo{person}{Aokun Chen}, \bibinfo{person}{Nima PourNejatian}, \bibinfo{person}{Hoo~Chang Shin}, \bibinfo{person}{Kaleb~E Smith}, \bibinfo{person}{Christopher Parisien}, \bibinfo{person}{Colin Compas}, \bibinfo{person}{Cheryl Martin}, \bibinfo{person}{Anthony~B Costa}, \bibinfo{person}{Mona~G Flores}, {et~al\mbox{.}}} \bibinfo{year}{2022}\natexlab{}.
\newblock \showarticletitle{A large language model for electronic health records}.
\newblock \bibinfo{journal}{\emph{NPJ Digital Medicine}} \bibinfo{volume}{5}, \bibinfo{number}{1} (\bibinfo{year}{2022}), \bibinfo{pages}{194}.
\newblock


\bibitem[Yeh et~al\mbox{.}({[n.\,d.]})]%
        {yeh2023toward}
\bibfield{author}{\bibinfo{person}{Chin-Chia~Michael Yeh}, \bibinfo{person}{Xin Dai}, \bibinfo{person}{Huiyuan Chen}, \bibinfo{person}{Yan Zheng}, \bibinfo{person}{Yujie Fan}, \bibinfo{person}{Audrey Der}, \bibinfo{person}{Vivian Lai}, \bibinfo{person}{Zhongfang Zhuang}, \bibinfo{person}{Junpeng Wang}, \bibinfo{person}{Liang Wang}, {et~al\mbox{.}}} \bibinfo{year}{[n.\,d.]}\natexlab{}.
\newblock \showarticletitle{Toward a foundation model for time series data}. In \bibinfo{booktitle}{\emph{Proceedings of the 32nd ACM International Conference on Information and Knowledge Management}}.
\newblock


\bibitem[Yu et~al\mbox{.}(2023)]%
        {yu2023temporal}
\bibfield{author}{\bibinfo{person}{Xinli Yu}, \bibinfo{person}{Zheng Chen}, \bibinfo{person}{Yuan Ling}, \bibinfo{person}{Shujing Dong}, \bibinfo{person}{Zongyi Liu}, {and} \bibinfo{person}{Yanbin Lu}.} \bibinfo{year}{2023}\natexlab{}.
\newblock \showarticletitle{Temporal Data Meets LLM--Explainable Financial Time Series Forecasting}.
\newblock \bibinfo{journal}{\emph{arXiv preprint arXiv:2306.11025}} (\bibinfo{year}{2023}).
\newblock


\bibitem[Yuan et~al\mbox{.}(2024)]%
        {yuan2024unist}
\bibfield{author}{\bibinfo{person}{Yuan Yuan}, \bibinfo{person}{Jingtao Ding}, \bibinfo{person}{Jie Feng}, \bibinfo{person}{Depeng Jin}, {and} \bibinfo{person}{Yong Li}.} \bibinfo{year}{2024}\natexlab{}.
\newblock \bibinfo{title}{UniST: A Prompt-Empowered Universal Model for Urban Spatio-Temporal Prediction}.
\newblock
\newblock
\showeprint[arxiv]{2402.11838}~[cs.LG]


\bibitem[Yuan et~al\mbox{.}(2023)]%
        {yuan2023spatio}
\bibfield{author}{\bibinfo{person}{Yuan Yuan}, \bibinfo{person}{Jingtao Ding}, \bibinfo{person}{Chenyang Shao}, \bibinfo{person}{Depeng Jin}, {and} \bibinfo{person}{Yong Li}.} \bibinfo{year}{2023}\natexlab{}.
\newblock \showarticletitle{Spatio-temporal Diffusion Point Processes}.
\newblock \bibinfo{journal}{\emph{arXiv preprint arXiv:2305.12403}} (\bibinfo{year}{2023}).
\newblock


\bibitem[Yue et~al\mbox{.}(2022)]%
        {yue2022ts2vec}
\bibfield{author}{\bibinfo{person}{Zhihan Yue}, \bibinfo{person}{Yujing Wang}, \bibinfo{person}{Juanyong Duan}, \bibinfo{person}{Tianmeng Yang}, \bibinfo{person}{Congrui Huang}, \bibinfo{person}{Yunhai Tong}, {and} \bibinfo{person}{Bixiong Xu}.} \bibinfo{year}{2022}\natexlab{}.
\newblock \showarticletitle{Ts2vec: Towards universal representation of time series}. In \bibinfo{booktitle}{\emph{Proceedings of the AAAI Conference on Artificial Intelligence}}, Vol.~\bibinfo{volume}{36}. \bibinfo{pages}{8980--8987}.
\newblock


\bibitem[Yun et~al\mbox{.}(2023)]%
        {yun2023imputation}
\bibfield{author}{\bibinfo{person}{Taeyoung Yun}, \bibinfo{person}{Haewon Jung}, {and} \bibinfo{person}{Jiwoo Son}.} \bibinfo{year}{2023}\natexlab{}.
\newblock \showarticletitle{Imputation as Inpainting: Diffusion models for SpatioTemporal Data Imputation}.
\newblock  (\bibinfo{year}{2023}).
\newblock


\bibitem[Zhang et~al\mbox{.}(2023)]%
        {zhang2023self}
\bibfield{author}{\bibinfo{person}{Kexin Zhang}, \bibinfo{person}{Qingsong Wen}, \bibinfo{person}{Chaoli Zhang}, \bibinfo{person}{Rongyao Cai}, \bibinfo{person}{Ming Jin}, \bibinfo{person}{Yong Liu}, \bibinfo{person}{James Zhang}, \bibinfo{person}{Yuxuan Liang}, \bibinfo{person}{Guansong Pang}, \bibinfo{person}{Dongjin Song}, {et~al\mbox{.}}} \bibinfo{year}{2023}\natexlab{}.
\newblock \showarticletitle{Self-supervised learning for time series analysis: Taxonomy, progress, and prospects}.
\newblock \bibinfo{journal}{\emph{arXiv preprint arXiv:2306.10125}} (\bibinfo{year}{2023}).
\newblock


\bibitem[Zhang et~al\mbox{.}(2024)]%
        {zhang2024large}
\bibfield{author}{\bibinfo{person}{Xiyuan Zhang}, \bibinfo{person}{Ranak~Roy Chowdhury}, \bibinfo{person}{Rajesh~K. Gupta}, {and} \bibinfo{person}{Jingbo Shang}.} \bibinfo{year}{2024}\natexlab{}.
\newblock \bibinfo{title}{Large Language Models for Time Series: A Survey}.
\newblock
\newblock
\showeprint[arxiv]{2402.01801}~[cs.LG]


\bibitem[Zhang et~al\mbox{.}({[n.\,d.]})]%
        {zhang2022self}
\bibfield{author}{\bibinfo{person}{Xiang Zhang}, \bibinfo{person}{Ziyuan Zhao}, \bibinfo{person}{Theodoros Tsiligkaridis}, {and} \bibinfo{person}{Marinka Zitnik}.} \bibinfo{year}{[n.\,d.]}\natexlab{}.
\newblock \showarticletitle{Self-supervised contrastive pre-training for time series via time-frequency consistency}.
\newblock \bibinfo{journal}{\emph{Advances in Neural Information Processing Systems}}  \bibinfo{volume}{35} (\bibinfo{year}{[n.\,d.]}).
\newblock


\bibitem[Zhang et~al\mbox{.}(2021)]%
        {zhang2021cloudrca}
\bibfield{author}{\bibinfo{person}{Yingying Zhang}, \bibinfo{person}{Zhengxiong Guan}, \bibinfo{person}{Huajie Qian}, \bibinfo{person}{Leili Xu}, \bibinfo{person}{Hengbo Liu}, \bibinfo{person}{Qingsong Wen}, \bibinfo{person}{Liang Sun}, \bibinfo{person}{Junwei Jiang}, \bibinfo{person}{Lunting Fan}, {and} \bibinfo{person}{Min Ke}.} \bibinfo{year}{2021}\natexlab{}.
\newblock \showarticletitle{{CloudRCA}: A root cause analysis framework for cloud computing platforms}. In \bibinfo{booktitle}{\emph{Proceedings of the 30th ACM International Conference on Information \& Knowledge Management}}. \bibinfo{pages}{4373--4382}.
\newblock


\bibitem[Zhao et~al\mbox{.}(2023)]%
        {zhao2023survey}
\bibfield{author}{\bibinfo{person}{Wayne~Xin Zhao}, \bibinfo{person}{Kun Zhou}, \bibinfo{person}{Junyi Li}, \bibinfo{person}{Tianyi Tang}, \bibinfo{person}{Xiaolei Wang}, \bibinfo{person}{Yupeng Hou}, \bibinfo{person}{Yingqian Min}, \bibinfo{person}{Beichen Zhang}, \bibinfo{person}{Junjie Zhang}, \bibinfo{person}{Zican Dong}, {et~al\mbox{.}}} \bibinfo{year}{2023}\natexlab{}.
\newblock \showarticletitle{A survey of large language models}.
\newblock \bibinfo{journal}{\emph{arXiv preprint arXiv:2303.18223}} (\bibinfo{year}{2023}).
\newblock


\bibitem[Zhou et~al\mbox{.}(2021)]%
        {zhou2021informer}
\bibfield{author}{\bibinfo{person}{Haoyi Zhou}, \bibinfo{person}{Shanghang Zhang}, \bibinfo{person}{Jieqi Peng}, \bibinfo{person}{Shuai Zhang}, \bibinfo{person}{Jianxin Li}, \bibinfo{person}{Hui Xiong}, {and} \bibinfo{person}{Wancai Zhang}.} \bibinfo{year}{2021}\natexlab{}.
\newblock \showarticletitle{Informer: Beyond efficient transformer for long sequence time-series forecasting}. In \bibinfo{booktitle}{\emph{Proceedings of the AAAI conference on artificial intelligence}}, Vol.~\bibinfo{volume}{35}. \bibinfo{pages}{11106--11115}.
\newblock


\bibitem[Zhou et~al\mbox{.}(2022)]%
        {zhou2022fedformer}
\bibfield{author}{\bibinfo{person}{Tian Zhou}, \bibinfo{person}{Ziqing Ma}, \bibinfo{person}{Qingsong Wen}, \bibinfo{person}{Xue Wang}, \bibinfo{person}{Liang Sun}, {and} \bibinfo{person}{Rong Jin}.} \bibinfo{year}{2022}\natexlab{}.
\newblock \showarticletitle{Fedformer: Frequency enhanced decomposed transformer for long-term series forecasting}. In \bibinfo{booktitle}{\emph{International conference on machine learning}}. \bibinfo{pages}{27268--27286}.
\newblock


\bibitem[Zhou et~al\mbox{.}(2023b)]%
        {zhou2023one}
\bibfield{author}{\bibinfo{person}{Tian Zhou}, \bibinfo{person}{Peisong Niu}, \bibinfo{person}{Xue Wang}, \bibinfo{person}{Liang Sun}, {and} \bibinfo{person}{Rong Jin}.} \bibinfo{year}{2023}\natexlab{b}.
\newblock \showarticletitle{One Fits All: Power General Time Series Analysis by Pretrained LM}.
\newblock \bibinfo{journal}{\emph{Advances in Neural Information Processing Systems}} (\bibinfo{year}{2023}).
\newblock


\bibitem[Zhou et~al\mbox{.}(2023a)]%
        {zhou2023maintain}
\bibfield{author}{\bibinfo{person}{Zhengyang Zhou}, \bibinfo{person}{Qihe Huang}, \bibinfo{person}{Kuo Yang}, \bibinfo{person}{Kun Wang}, \bibinfo{person}{Xu Wang}, \bibinfo{person}{Yudong Zhang}, \bibinfo{person}{Yuxuan Liang}, {and} \bibinfo{person}{Yang Wang}.} \bibinfo{year}{2023}\natexlab{a}.
\newblock \showarticletitle{Maintaining the Status Quo: Capturing Invariant Relations for OOD Spatiotemporal Learning}. In \bibinfo{booktitle}{\emph{Proceedings of the 29th ACM SIGKDD Conference on Knowledge Discovery and Data Mining}} \emph{(\bibinfo{series}{KDD '23})}. \bibinfo{pages}{3603–3614}.
\newblock
\showISBNx{9798400701030}


\bibitem[Zhu et~al\mbox{.}(2024)]%
        {zhu2024difftraj}
\bibfield{author}{\bibinfo{person}{Yuanshao Zhu}, \bibinfo{person}{Yongchao Ye}, \bibinfo{person}{Shiyao Zhang}, \bibinfo{person}{Xiangyu Zhao}, {and} \bibinfo{person}{James Yu}.} \bibinfo{year}{2024}\natexlab{}.
\newblock \showarticletitle{Difftraj: Generating gps trajectory with diffusion probabilistic model}.
\newblock \bibinfo{journal}{\emph{Advances in Neural Information Processing Systems}}  \bibinfo{volume}{36} (\bibinfo{year}{2024}).
\newblock


\bibitem[Zhu et~al\mbox{.}(2023)]%
        {zhu2023energy}
\bibfield{author}{\bibinfo{person}{Zhaoyang Zhu}, \bibinfo{person}{Weiqi Chen}, \bibinfo{person}{Rui Xia}, \bibinfo{person}{Tian Zhou}, \bibinfo{person}{Peisong Niu}, \bibinfo{person}{Bingqing Peng}, \bibinfo{person}{Wenwei Wang}, \bibinfo{person}{Hengbo Liu}, \bibinfo{person}{Ziqing Ma}, \bibinfo{person}{Xinyue Gu}, {et~al\mbox{.}}} \bibinfo{year}{2023}\natexlab{}.
\newblock \showarticletitle{Energy forecasting with robust, flexible, and explainable machine learning algorithms}.
\newblock \bibinfo{journal}{\emph{AI Magazine}} \bibinfo{volume}{44}, \bibinfo{number}{4} (\bibinfo{year}{2023}), \bibinfo{pages}{377--393}.
\newblock


\end{thebibliography}
